%% file: aistats2025.tex
\documentclass[twoside]{article}

\usepackage[accepted]{aistats2025}
\usepackage{microtype}
\usepackage{graphicx}
\usepackage{subfigure}
\usepackage{booktabs} 
\usepackage{enumitem} 

\usepackage{svg}
\usepackage{tikz}
\usepackage{amsmath}
\usepackage{amssymb}

\usepackage{algorithmic}
\usepackage{mathtools}
\usepackage{amsthm}
\usepackage{pythonhighlight}
\usepackage[toc,page]{appendix}
\usepackage[ruled]{algorithm2e}
\theoremstyle{plain}
\newtheorem{theorem}{Theorem}[section]

\newtheorem{lemma}[theorem]{Lemma}

\newtheorem{assumption}[theorem]{Assumption}
\theoremstyle{definition}
\newtheorem{definition}[theorem]{Definition}
\theoremstyle{remark}
\newtheorem{remark}[theorem]{Remark}

\usepackage[utf8]{inputenc} 
\usepackage[T1]{fontenc}    
\usepackage{hyperref}       
\usepackage{url}            
\usepackage{booktabs}       
\usepackage{amsfonts}       
\usepackage{nicefrac}       
\usepackage{microtype}      
\usepackage{xcolor}         

\usepackage{multirow}
\usepackage{wrapfig}
\usepackage{graphicx}

\usepackage{wrapfig}
\usepackage{booktabs}
\usepackage{amsmath,amssymb,amsfonts}
\usepackage{amsthm}
\usepackage{graphicx}
\usepackage{multirow}
\usepackage{pifont}
\usepackage{graphicx}
\usepackage{subcaption}
\usepackage{booktabs}
\usepackage{hyperref}
\usepackage{amsmath}

\usepackage{amssymb}
\usepackage{amsfonts}
\usepackage{mathtools}
\usepackage{mathrsfs}
\usepackage{wrapfig}
\usepackage{stfloats}
\usepackage{xcolor}
\usepackage[capitalize,noabbrev]{cleveref}
\usepackage{textcomp}
\usepackage{xspace}
\usepackage{makecell}
\usepackage{threeparttable}
\usepackage{tablefootnote}
\usepackage{footnote}
\usepackage[hang,flushmargin]{footmisc}
\usepackage{balance}
\usepackage{comment}
\usepackage[normalem]{ulem}
\usepackage{cellspace}
\usepackage{array, tabularx, boldline}

\usepackage{multirow}

\newcommand{\B}{\boldsymbol}

\newcommand{\FF}{\mathcal{F}}

\newcommand{\mbR}{\mathbf{R}} %

\newcommand{\tabincell}[2]{\begin{tabular}{@{}#1@{}}#2\end{tabular}}

\newcommand{\cred}{\textcolor{red}}
\newcommand{\yes}{\textcolor{teal}{\ding{51}}}
\newcommand{\no}{\cred{\ding{55}}}

\usepackage[utf8]{inputenc} 
\usepackage[T1]{fontenc}    
\usepackage{hyperref}       
\usepackage{url}            
\usepackage{booktabs}       
\usepackage{amsfonts}       
\usepackage{nicefrac}       
\usepackage{microtype}      
\usepackage{xcolor}         
\begin{document}

%
\runningtitle{Prior-Fitted Networks Scale to Larger Datasets  When Treated as Weak Learners}

%
\runningauthor{Yuxin Wang, Botian Jiang,  Yiran Guo, Quan Gan,  David Wipf, Xuanjing Huang, Xipeng Qiu}

\twocolumn[
\aistatstitle{Prior-Fitted Networks Scale to Larger Datasets  \\ When Treated as Weak Learners}
\aistatsauthor{Yuxin Wang$^{1,2}$\And Botian Jiang$^{1}$\And Yiran Guo$^{1}$}
\aistatsauthor{Quan Gan$^{3}$\And David Wipf$^{3}$\And Xuanjing Huang$^{1,2}$ \And Xipeng Qiu$^{1}$ }

\aistatsaddress{$^{1}$School of Computer Science, Fudan University}
\vspace{-8mm}
\aistatsaddress{$^{2}$Institute of Modern Languages and Linguistics, Fudan University}
\vspace{-8mm}
\aistatsaddress{$^{3}$Amazon}]

\begin{abstract}

Prior-Fitted Networks (PFNs) have recently been proposed to efficiently perform tabular classification tasks. Although they achieve good performance on small datasets, they encounter limitations with larger datasets. These limitations include significant memory consumption and increased computational complexity, primarily due to the impracticality of incorporating all training samples as inputs within these networks. To address these challenges, we investigate the fitting assumption for PFNs and input samples. Building on this understanding, we propose \textit{BoostPFN} designed to enhance the performance of these networks, especially for large-scale datasets. We also theoretically validate the convergence of BoostPFN and our empirical results demonstrate that the BoostPFN method can outperform standard PFNs with the same size of training samples in large datasets and achieve a significant acceleration in training times compared to other established baselines in the field, including widely-used Gradient Boosting Decision Trees (GBDTs), deep learning methods and AutoML systems. High performance is maintained for up to 50x of the pre-training size of PFNs, substantially extending the limit of training samples. Through this work, we address the challenges of efficiently handling large datasets via PFN-based models, paving the way for faster and more effective tabular data classification training and prediction process. Code is available at \href{Github}{https://github.com/yxzwang/BoostPFN}.
\end{abstract}

\vspace*{-0.2cm}
\section{Introduction}
\label{sec:intro}
\vspace*{-0.2cm}

\footnote{This work was completed when the first author was during an internship at the AWS Shanghai AI Lab.}Tabular datasets are increasingly garnering attention in various practical applications across fields such as finance~\cite{clements2020sequential}, healthcare~\cite{esteva2019guide}, and scientific research~\cite{deiana2022applications}, among others~\cite{benjelloun2020google,tang2020customer,ulmer2020trust,urban2021deep}. As we continue to amass larger and more complex datasets, the need for efficient and effective methods of data analysis becomes ever more critical. Traditional approaches often fall short due to prohibitively long training times when dealing with large datasets. This limitation not only hampers the speed of data processing but also impacts the overall feasibility of utilizing large-scale tabular data in time-sensitive scenarios~\cite{deiana2022applications}.

Recently, there has been a significant shift towards exploring novel methodologies that can overcome the limitations of traditional data processing techniques. One such promising development is the advent of the in-context learning method known as Prior-fitted Networks (PFNs)~\cite{muller2021transformers}. This innovative approach is capable of making predictions without the need for training on the training set, instead utilizing training data points as input tokens to the model architecture, thereby offering a much faster training and prediction process compared to conventional methods. The outstanding model in Prior-fitted Networks: TabPFN~\cite{hollmann2023tabpfn}, uses a Transformer as the model architecture. Therefore, the primary issue with TabPFN on large datasets lies in the scalability of the Transformer: when applied to large datasets, it becomes impractical due to excessive memory consumption and computational complexity. The computational complexity of the Transformer increase quadratically with input length, which, for TabPFN, corresponds to the size of the training dataset. Despite these challenges, TabPFN's ability to rapidly process large tabular datasets remains a compelling feature, urging us to find a way to apply it to large datasets.

\begin{table}[]
    \centering
    \caption{Comparison of BoostPFN with GBDTs, Deep Learning (D.L.) methods and TabPFN. }
\resizebox{0.5\textwidth}{!}{\begin{tabular}{c|ccc}
    \toprule
          Models & \tabincell{l}{Strong Performance with \\Limited Training Samples}  & Training Efficiency & Large Datasets\\
          \midrule
         GBDTs & \no & \no & \yes \\
         D.L. & \no & \no & \yes \\
         TabPFN & \yes & \yes & \no \\
         \midrule
         BoostPFN & \yes & \yes & \yes \\
    \bottomrule
    \end{tabular}}
    \label{tab:tictable}
\end{table}

\begin{figure*}[htbp]
    \centering
    \includegraphics[width=0.8\textwidth]{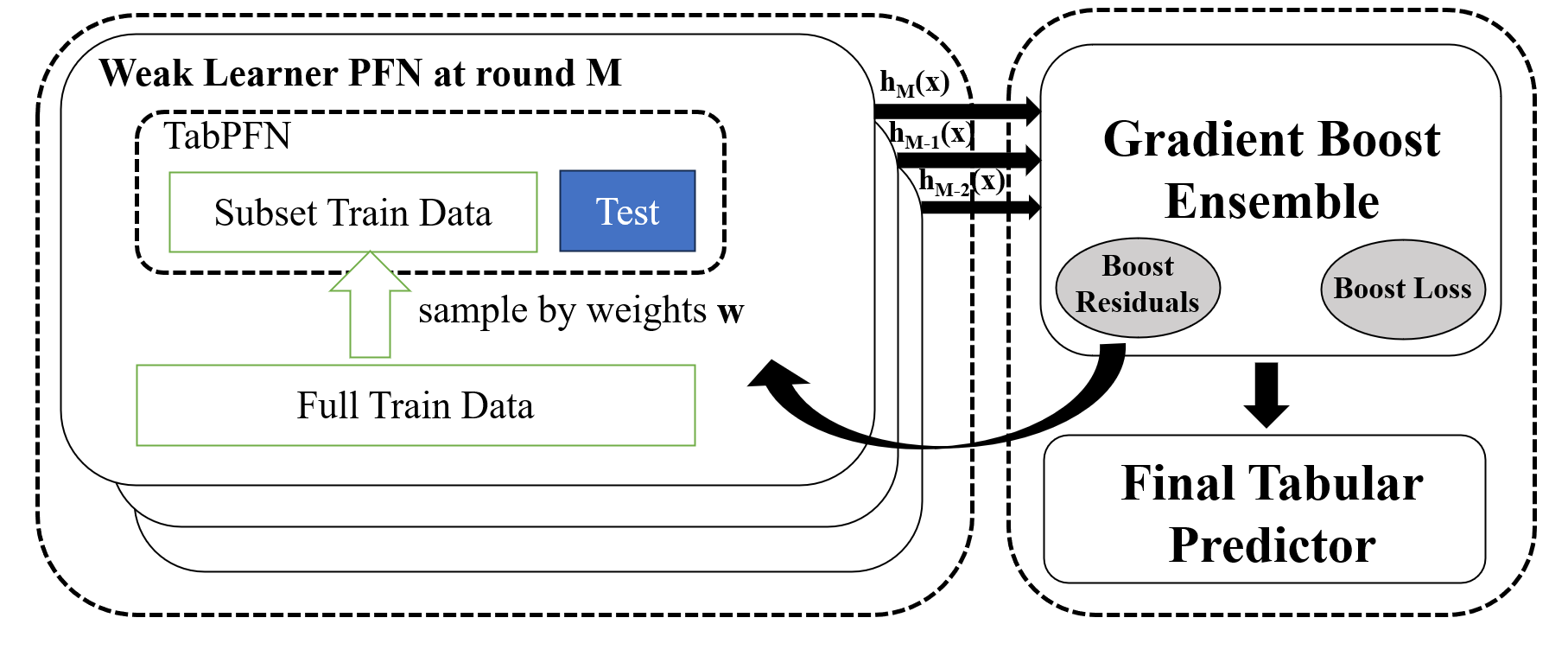}
    \caption{Overview of BoostPFN. Each PFN with a subset of train data is viewed as a weak learner $h_{M}(x)$ at round $M$. While adding new weak learners into the ensemble, new weak learner is governed by updating new sampling weights with the boosting residuals from the existing ensemble.  }
    \label{fig:GboostArchitecture}

\end{figure*}

In this paper, we introduce a way to treat PFN as a type of "weak" learner in the gradient boosting process, achieved through the Fitting Assumption \ref{prop:fittingassumption}. We then propose BoostPFN as an extension of Gradient Boosting~\cite{friedman2001greedy} for ensembling these PFNs, as illustrated in Figure \ref{fig:GboostArchitecture}. We propose three updating methods based on the Fitting Assumption, including the AdaBoost updating, which allows our method to be viewed as a natural extension of the well-known boosting method AdaBoost~\cite{hastie2009multi}. These methods explore how ensemble learning techniques can be integrated with PFNs to create a more robust prediction model. We delve into both the theoretical and empirical validation of BoostPFN and demonstrate that our proposed methods not only achieve high performance but also maintain the efficiency in training that is critical for handling large-scale tabular datasets, compared to GBDTs and other deep learning methods that are trained explicitly. We compare BoostPFN with other types of models in Table \ref{tab:tictable}. Our contributions are as follows:

\begin{itemize} \item TabPFN is designed for fast training on tabular classification tasks but cannot handle large datasets, while BoostPFN extends the training dataset size and retains high performance, accommodating up to \textit{50x} the pre-training size of TabPFN. Importantly, even when limited by TabPFN's pre-training size of 1024, which struggles to yield relatively good results compared to other models, our method paves the way for extending any new or improved PFN with a larger pre-training size and better training data. \item We delve into the theoretical analysis of BoostPFN, showing that it is a type of Randomized Gradient Boosting Machine with a convergence guarantee. Empirical results also demonstrate the convergence of boosting loss on both the training and test sets when overfitting does not occur. \item We conduct extensive experiments on datasets of varying sizes, from small to large, to assess the performance of BoostPFN, highlighting its efficacy and scalability. \end{itemize}

\vspace*{-3mm}
\section{Preliminaries}
\vspace*{-3mm}
\label{sec:preliminary}
\textbf{Posterior Predictive Distribution.} The Posterior Predictive Distribution (PPD) is a concept in Bayesian statistics that combines the information from both the observed data and the posterior distribution of the model parameters. It is a predictive distribution for future observations based on the updated beliefs about the model parameters after incorporating the observed data.
\begin{equation}
    p(y|x,D) \propto \int_{\Phi} p(y|x,\phi) p(D|\phi) p(\phi) d\phi.
    \label{eq:ppd}
\end{equation}

\textbf{PFNs and Architecture.} PFNs are proposed in~\cite{muller2021transformers,hollmann2023tabpfn} to approximate the Posterior Predictive Distribution (PPD) for supervised learning. For a test sample $x_{test}$ and train datasets $D:={(x_i, y_i)}_{i \in \{1,\dots,n\}}$, the prediction of TabPFN~\cite{hollmann2023tabpfn} can be written as 
\begin{align}
\label{eq:tabpfn_implementation}
    p(\hat{y}_{test})=q_{\theta}(\hat{y}_{test}|x_{text},D_{train}),
\end{align}
where $\theta$ are the parameters of TabPFN. The architecture of TabPFN is a Transformer without positional embeddings. Training samples and test samples are concatenated together and taken as the input of Transformer. Probabilities for each class are given by Transformer as output.

\textbf{Gradient Boosting.} Gradient Boosting~\cite{friedman2001greedy} is a powerful machine learning technique that constructs a predictive model $F_m(x)$ in the form of an ensemble of weak predictive models (commonly decision trees) at iteration $m$, 
\vspace*{-0.2cm}
\begin{align}
\label{eq:gb_addition}
     F_m(x) = F_{m-1}(x) + \gamma_m h_m(x),
\end{align}
where $h_m(x)$ is a weak learner selected from the candidate learner space $\mathcal{H}$ and $\gamma_m$ is the learning rate learned by line search. The $h_m(x)$ is selected to fit the residual errors $r_{m} = -\left[\frac{\partial \ell(y, F_{m-1}(x))}{\partial F_{m-1}(x_i)}\right]$ of the prior models,
\vspace*{-0.3cm}
\begin{align}
\label{eq:gb_fitting_residual}
    h_m=\text{args}\min_{h \in \mathcal{H}}\sum_{i=1}^{|D_{train}|}||r_{m}[i]-h(x_i)||_2^2,
\end{align}
where $r_m[i]$ is the $i$th value of vector $r_m$.
\vspace*{-0.1cm}

\section{PFN Scalability Challenges}
\vspace*{-0.1cm}
\label{sec:scalabilitychallenges}
\textbf{Long-range Transformers.} While the Transformer is the backbone of TabPFN, it faces challenges with large datasets because TabPFN considers all training samples at once. There are Transformer variants~\cite{guo2019starTransformer,child2019generating,beltagy2020longformer,zaheer2021big,pmlr-v162-wang22l} that enable long-range inputs and show promising results compared to the vanilla Transformer with thousands of input tokens. However, for \textit{larger} datasets with more training samples, the capability of handling thousands of input tokens is merely a drop in the bucket, and pre-training inevitably requires substantial time. Although one direction is to develop more efficient Transformer architectures, we focus on the alternative direction: applying a trained PFN to larger datasets.

\textbf{Sampling Methods.} A straightforward approach is to sample a small portion of training samples as the input training datasets, which is common for Transformers. In Natural Language Processing, training datasets are retrieved by a Retriever~\cite{rajaraman_ullman_2011,robertson2009probabilistic,nogueira2019document,karpukhin2020dense,zhang2020dcbert,khattab2020colbert,izacard2022unsupervised,xu2022laprador,shi2023replug,su2023embedder,li2023llatrieval} that considers the relationship between the test sample and candidate samples. However, retriever-based methods retrieve training samples differently for each test sample, making implementation challenging when the number of test samples is large. Additionally, retrieving training samples from a vast number of candidates is time-consuming, and results are not guaranteed because existing retrievers can only accept thousands of candidates. Therefore, we opt to use simple sampling for training sample selection. However, this simple approach is very heuristic and may not yield good performance. To improve performance, we can employ ensemble methods. One of the simplest ensemble methods is Bagging~\cite{breiman1996bagging}, which samples multiple times and averages the predictions. We include this as our ensemble baseline method. For better performance, we will next introduce our method of scaling PFNs through boosting.

\vspace*{-0.2cm}
\section{Scaling PFNs through Boosting}
\vspace*{-0.2cm}

    We propose a gradient boosting method for PFNs called BoostPFN. In the normal procedure of gradient boosting discussed in Section \ref{sec:preliminary}, an essential step is to fit a new weak learner based on training residuals, as shown in Eq. \ref{eq:gb_fitting_residual}. For traditional weak learners, such as decision trees, this is easy to implement. However, for PFNs, there is no existing method to fit a new PFN while keeping the parameters of the Transformer architecture fixed. Optimizing the parameters of a PFN using gradient methods is time-consuming and undermines the primary advantage of PFNs, which is their negligible training time. To maintain \textit{training efficiency} and enable the fitting step, we first propose learnable sampling optimization for PFNs. Since the input datasets are a crucial part of predicting the target sample, optimizing the input dataset can also enhance prediction accuracy. 
   For normalized sampling weights $w$, a specified PFN $q_{\theta}$, and a test sample $x_{test}$, a PFN with a sampled input dataset is defined as
\vspace*{-0.1cm}
\begin{align}
    q_{(\theta,w)}(y|x_{test},D)=q_{\theta}(y|x_{test},D_w^z),
\end{align}
where $D$ is the full training set and $D_w^z$ is the sampled input dataset generated by weights $w \in \mbR^{|D|}$ and sampling size $z \in (0,|D|]$ which determines the sampled training set size $|D_w^z|=z$.

\vspace*{-3mm}
\subsection{Optimizing Input Datasets}

We first describe how we optimize the input dataset by updating $w$ using pseudo-residuals, similar to the approach in gradient boosting.

\begin{assumption}
\label{prop:fittingassumption}
    Given a sample $(x_i,y_i)$, and any training set $\tilde{D}$, the probability of predicting the right class $y_i$ on $x_i$ be
    \begin{align}
        &q_{\theta}(y_i|x_i,\tilde{D}\cup \{(x_i,y_i)\} ) > 
        \\ \nonumber &\max \Big[q_{\theta}(y_i|x_i,\tilde{D}) , q_{\theta}(y_i|x_i,\tilde{D} \cup \{(x_j,y_j)\})\Big],
    \end{align}
\end{assumption}

Assumption \ref{prop:fittingassumption} can be explained simply: adding the target sample or replacing one sample in the training set with the target sample will improve prediction performance. This seems straightforward for a GP (Gaussian Process)~\cite{seeger2004gaussian}, as fitting a GP to a single observation sharpens the posterior predictive distribution as it increases. However, PFNs are implemented in practice using a pretrained Transformer architecture, which does not inherently behave like a GP, and no such experiments have been conducted before. Therefore, we also perform experiments to empirically validate this assumption, as described in Section \ref{sec:addsampleablation}.

With this assumption, we can propose a simple idea: increasing the sampling weight $w[i]$ for $D[i]$ improves the probability of sampling that target $i$, thereby increasing the expectation of correctly predicting the label. Although this is a complex condition in practice, we can analyze it in a simplified scenario, as shown in Section \ref{appendix:proofforlemma}. This motivates a straightforward approach for updating sampling weights in a given dataset. In the next section, we explain in detail how to update them in practice.
\subsection{Updating Sampling Weights}
\label{sec:updatingmethods}
In this section, we discuss how to update sampling weights. Like the fitting residuals of gradient boosting in Eq. \ref{eq:gb_fitting_residual}, we also want our our updating methods to depend on the residuals $r_{m} = -\left[\frac{\partial \ell(y, F_{m-1}(x))}{\partial F_{m-1}(x_i)}\right]$. However, there is no previous work on how to apply residuals in sampling weights updating. Starting by remembering that in the gradient descent, the absolute value of gradient goes to zero when the loss come to the optimal plain. We come up with a rule for updating sampling weights, that is to give the sample with larger absolute residual more weights in the next round. Then we propose three updating methods in this section.   

\textbf{ExpHadamard Updating.} New weights are the hadamard products of this round's weights and the exponential of the residual, 
\vspace*{-0.2cm}
\begin{align}
\label{eq:exphadamardupdating}
w_m=w_{m-1} \odot \text{exp}(|r_m|).
\end{align}
\textbf{Hadamard Updating.} New weights are the hadamard products of this round's weights and the residual, 
\vspace*{-0.2cm}
\begin{align}
\label{eq:hadamardupdating}
w_m=w_{m-1} \odot |r_m|.
\end{align}
\textbf{AdaBoost Updating.} Also, we adopt the updating in AdaBoost~\cite{hastie2009multi} given by
\begin{align}
    \label{eq:adaboostupdating}
     \epsilon_{m}&= w_{m-1} \otimes \textbf{1}\{h_{m-1}(X[i])\neq Y[i]\} \nonumber \\
     \alpha_m&=\text{log}\left(\frac{1-\epsilon_m}{\epsilon_m}\right) + \text{log}(K-1),\\
        w_m&=w_{m-1} \odot \text{exp}\left(\alpha_m\textbf{1}\{h_{m-1}(X[i])\neq Y[i]\}\right), \nonumber
\end{align}

where $\epsilon_{m}$ is the summation of the sampling weights of samples that the last weak learner predicts wrong on and $K$ is the class number of targets. And we increase the weights for wrong samples and decrease the weights for right samples based on $\epsilon_{m}$. As an alternative, the $\gamma_m$ could also be replaced by $\alpha_m$ as in AdaBoost. In practice, we choose the $\gamma_m$ with better performances. Note that although there is no residuals in the AdaBoost Updating, $\epsilon_{m}$ can be seen as a kind of residual for a specified AdaBoost loss funciton $\ell_{AdaBoost}$~\cite{hastie2009elements}.

In practice, we view the updating method as a hyperparameter of BoostPFN and choose the best one for each dataset. The ablation over different updating methods are shown in Section \ref{sec:ablation-updatingmethods}. 
\vspace*{-2mm}
\subsection{BoostPFN}
\vspace*{-2mm}

By combining the ingredients from above, with the the gradient boosting process discussed in Section \ref{sec:preliminary}, we arrive at our BoostPFN approach, the steps of which are summarized in Algorithm \ref{alg:BoostPFN}. 

\begin{algorithm}[]
\caption{BoostPFN}
\label{alg:BoostPFN}
\begin{algorithmic}[1]
\REQUIRE Training data $D=(X,Y)$, PFN model $q_{\theta}(y|x,D)$ , sampling weights $w$ for training samples, sampling size $z$, boosting round number $M$ and boosting loss $\ell$. \\ 
\hspace*{-0.6cm}\KwOut{BoostPFN predictor $F(x)$.}
\STATE Initialize by $F_0=0$.
\FOR {$m$ = 1 to $M$}

    \STATE Compute residual: $r_m=-\frac{\partial \ell(y,F_{m-1}(X))}{\partial F_{m-1}(X)}$.
    \\
    \STATE Update sampling weights $w_m  = Update(w_{m-1},r_m)$ via methods in Section \ref{sec:updatingmethods}.  
    
    \STATE $w_m  \leftarrow  \text{\textbf{Normalize}}(w_m)$.
    
    \STATE Do non-replacement sampling with $w_m$ to obtain a subset of training data $D_{w_m}^z$ 

    \STATE Select new weak learner $h_m(x)$ as the PFN model $q_{\theta}(y|x,D_{w_m}^z)$.  \\
    
    \STATE Line search for $\gamma_m$ by $\gamma_m=\arg \text{min}_{\gamma} \sum_{i=1}^{N_t} \ell(Y[i], F_{m-1}(X[i]) + \gamma h_m(X[i])$ .
    \STATE Get new predictor $F_m(x)=F_{m-1}(x) + \gamma_m h_m(x).$

\ENDFOR
\STATE \textbf{return} $F(x)=F_M(x)$ \\
\end{algorithmic}
\end{algorithm}

\vspace*{-1mm}
\section{Analysis of BoostPFN}
\vspace*{-1mm}
\subsection{Time Complexity}
\vspace*{-1mm}
\begin{table}[htbp]
    \centering
    \caption{Time Complexity for PFNs. We use $T_i$ to denote the inference time for TabPFN. $M$ is the round we use for bagging or boosting. $T_B$ is the time cost of boosting operations in each boosting round.}
    \begin{tabular}{c|cc}
    \toprule
         TabPFN & Bagging & BoostPFN \\
         \midrule
          $T_i$ & $MT_i$ & $M(T_i+T_B)$ \\
          \bottomrule
    \end{tabular}
    
    \label{tab:complexityforpfns}
\end{table}
We briefly show the time complexity in Table \ref{tab:complexityforpfns}. We consider the complexity analysis in a scenario with limited resources. Specifically, given the hardware available, we cannot load the entire test set due to out-of-memory (OOM) issues. Note that even with a small number of training samples, the OOM problem can still occur if the test set is large enough. Therefore, in practice, we have to split the test set into $N_{test}$ batches, and the inference of one batch consumes all the available memory (which is possible for millions of test samples). In this scenerio, the time complexity of BoostPFN is close to the simple Bagging method except for the addtion of inevitable boosting operations, which is an acceptable solution for large datasets. Then we describe in detail those symbols.

\textbf{TabPFN.}  One inference of TabPFN on a single batch costs $O(L^2)$ time, where $L = |D_m| + B$ represents the input length, equal to the sum of the length of the sampled dataset $|D_m|$ and the batch size $B$. For one inference of TabPFN on the entire test set, we need $T_i$ time. If we consider the sampling time negligible, this is also the time complexity for sampling a training set and performing TabPFN inference on the whole test set, which can be further reduced to
\begin{align}
     T_i&=O(N_{test}(|D_m|+B)^2)\\\nonumber
    &=O(N_{test}(|D_m|^2+B^2+2|D_m|B))\\\nonumber
    &=O\left(\frac{|D_m|^2}{B}|D_{test}|+B|D_{test}|+2|D_m||D_{test}|)\right)\\\nonumber
    &=O\left(\left(\frac{|D_m|^2}{B}+B+2|D_m|\right)|D_{test}|\right),
\end{align}
where we use the fact that $O(N_{test}B)=O(|D_{test}|)$.

\textbf{Bagging.} As an easy ensemble method mentioned in Section \ref{sec:scalabilitychallenges}, bagging is generally implemented in parallel. However, in our scenario, parallel inference on the test set is impossible because one inference on a single batch consumes all the GPU memory available. Thus, if we take $M$ bags and ignore the sampling time, the time complexity for bagging with TabPFN is $MT_i$.

\textbf{BoostPFN.} For the time-consuming sampling gradient boosting process, we observe that all operations, including updating weights or performing line searches for coefficients, are linear with respect to the training set. Therefore, the total time complexity for these operations is $T_B=O(T|D_{test}|)$, where $T$ is a constant. For $M$ rounds of boosting, the time complexity is $M(T_i+T_B)$, which is very close to the time complexity of the bagging method. Our analysis shows that, in a limited-resource scenario, BoostPFN can achieve similar time efficiency to bagging.
\vspace*{-1mm}
\subsection{Convergence}
\vspace*{-1mm}
We next examine the convergence of Algorithm \ref{alg:BoostPFN}, because methods that ensure convergence are typically more dependable, reproducible, and possibly efficient. We first define $F^*(x)$ as the optimal predictor that minimizes the loss $\ell$:
\begin{align}
    F^*(x)=\arg\min_{F(x)\in \mathcal{F}} \ell(y,F(x)),
\end{align}
where $\mathcal{F}$ is the searching space of possible predictors as determined by the associated family of
weak learners $\mathcal{H}$. Then we show the convergence in Theorem \ref{theorem:RGBM}.
\begin{theorem}
\label{theorem:RGBM}
Let $F_M(x)$ denote the BoostPFN predictor produced by Algorithm \ref{alg:BoostPFN} after $M$ steps. Then if the loss $\ell$ is $\sigma$-smooth and has a bounded level set, we have that 
\vspace*{-0.2cm}
\begin{align}
\label{eq:theorem}
    \mathbb{E} \big[|\ell(y,F_M(x))-\ell(y,F^*(x))|\big] \leq O\left(\frac{\sigma}{M}\right).
\end{align}
\end{theorem}
\vspace*{-0.2cm}
while $\sigma$-smooth is defined in 
\begin{definition}
 $\ell$ is $\sigma$-smooth if for any $y$ and predictions $f_1$ and $f_2$, it holds that
$$
\ell(y, f_1) \le \ell(y, f_2) + \frac{\partial \ell (y, f_2)}{\partial f}(f_1-f_2) + \frac{\sigma}{2} (f_1-f_2)^2 .
$$

\end{definition}
 The proof of Theorem \ref{theorem:RGBM} is deferred to Appendix \ref{appendix:prooffortheorem}, which is stratified by first proving that BoostPFN is a kind of RGBM (Randomized Gradient Boosting Machine)~\cite{lu2020randomized} and then using some known properties of RGBM. While our loss $\ell$ for classification is the same as RGBM in ~\cite{lu2020randomized} that fits the criterion, Theorem \ref{theorem:RGBM} validates the convergence of BoostPFN. We also empirically validate the convergence of BoostPFN in Section \ref{sec:convergenceemperical}.
 \vspace*{-1mm}
\section{Experiments}
\vspace*{-1mm}
We compare BoostPFN with other tabular baselines on datasets from small to large to show its effectiveness and training efficiency.

\textbf{Datasets and Baselines.} We use the same 30 datasets in~\cite{hollmann2023tabpfn} from OpenML Benchmarking Suites~\cite{bischl2021openml} and 30 larger datasets including real-world datasets and artificial datasets generated by Bayesian Neural Networks with more than 100 thousand samples and less than 100 features from OpenML Benchmark. Full datasets are listed in Appendix \ref{appendix:datasets-statistics}. We split the datasets following \cite{hollmann2023tabpfn} in 50/50 for training and test in 5 random seeds and report the average results. For larger datasets in Table \ref{tab:largetest_datasets_table}, we also conduct experiments on a limited size of training samples (5000 and 50000 samples respectively). For baselines, we use LightGBM~\cite{ke2017lightgbm}, XGBoost~\cite{chen2016xgboost}, CatBoost~\cite{prokhorenkova2018catboost} for GBDTs, and FT-Transformer~\cite{gorishniy2021revisiting}, SAINT~\cite{somepalli2021saint} for deep learning models, and an AutoML system AutoGluon~\cite{erickson2020autogluon}. We include TabPFN as a candidate baseline when not encountering OOM (out of memory) problems. We also include the Bagging method for TabPFN as an ensemble baseline. To get different number of training samples, we sample from the whole training set randomly. And the number of weak learners in ensemble model is 10 for 5,000 training samples, 100 for 50,000 and 1,000 for full training set. 

 \textbf{Implementations.} For evaluation metrics, we follow~\cite{hollmann2023tabpfn} and use AUC-OVO (which is the results of AUC by one class versus one class), which is most important in tabular datas because of unbalanced class samples. For boosting and bagging methods, ensemble numbers are fixed to 10 if not mentioned specifically. The input data is limited to 500 samples considering the balance of performance and time consumption. Full experimental implementation can be found in Appendix \ref{appendix:experimentalimplementation}.
\begin{table*}[htbp]
    \centering

    \small
        \caption{Mean AUCs on large tabular datasets with different number of training samples. The best result is \textbf{bolded} and the second best one is \underline{underlined} in each row. Results for each dataset are shown in Appendix \ref{appendix:fullexperiments}. }
    \label{tab:resultsonlargedatasets}
   \resizebox{\textwidth}{!}{ \begin{tabular}{l|lllllll|lll}
\toprule
\# of Training &        LightGBM &        CatBoost &          XGBoost &               AutoGluon & FT-Trans.& SAINT &  TabPFN &           Bagging &           BoostPFN   \\
\midrule
 < 5000  & 0.884  &        0.891 &          0.890 &                        \textbf{0.895} & 0.875 & 0.824 &  \underline{0.894} &           0.893 &           \textbf{0.895}  \\
5,000  & 0.865  &        0.865 &          0.849 &                        0.844 & \underline{0.900}& 0.878&  0.885 &           0.874 &           \textbf{0.908}  \\
50,000  &0.883 &        0.864 &         0.855 &                   0.865& \textbf{0.910}& \underline{0.907} &  OOM &           0.888 &           \textbf{0.910} \\
> 50,000   & 0.915 &        0.911 &          \underline{0.916} &               0.915 & \textbf{0.918} & \textbf{0.918} &  OOM &           0.895 &              0.913 \\
\bottomrule
\end{tabular}}

\end{table*}

\textbf{Results.} All model performances are shown in Table \ref{tab:resultsonlargedatasets}. "<5000" shows the average of small datasets. Other rows represent the average results of large datasets with different numbers of training samples. Result for each dataset is listed in Appendix \ref{appendix:fullexperiments}. It is showed that BoostPFN has strong merits when the number of training samples is under 50000. When on small datasets whose number of training samples is lower than 5000, BoostPFN can get state-of-the-art AUC results together with AutoGluon, also better than TabPFN. When it comes to large datasets, BoostPFN can get state-of-the-art AUC results when the number of training samples is not higher than 50000,  while TabPFN meets OOM problems when it comes to 50,000 training samples. This demonstrates the effectiveness of gradient boosting when training set size is larger than pre-training size of TabPFN (which is 1024). Baseline models can get better performances when the training set size increases (except CatBoost with a bit lower with 50,000 training samples, maybe because of not well-tuned hyperparameters due to long training time and time limitation). When using training samples larger than 50000, BoostPFN cannot get very good results compared to baseline models (while still better than CatBoost) because the existing prior is trained only on 1,000 samples and up to 1,000 weak learners can quickly saturate the performance improvement. The experimental results show BoostPFN can extend the dataset size of pre-training for TabPFN to 50x and still remain high performances.

\begin{figure*}[htbp]
    \centering
    \caption{We show Mean AUC, Wins AUC and Mean Rank. as a function of the time allowed to train and tune methods, on large datasets with training number equals to 5,000 from OpenML benchmarks.}
\begin{minipage}[t]{0.32\textwidth}
\centering
\includegraphics[width=\textwidth]{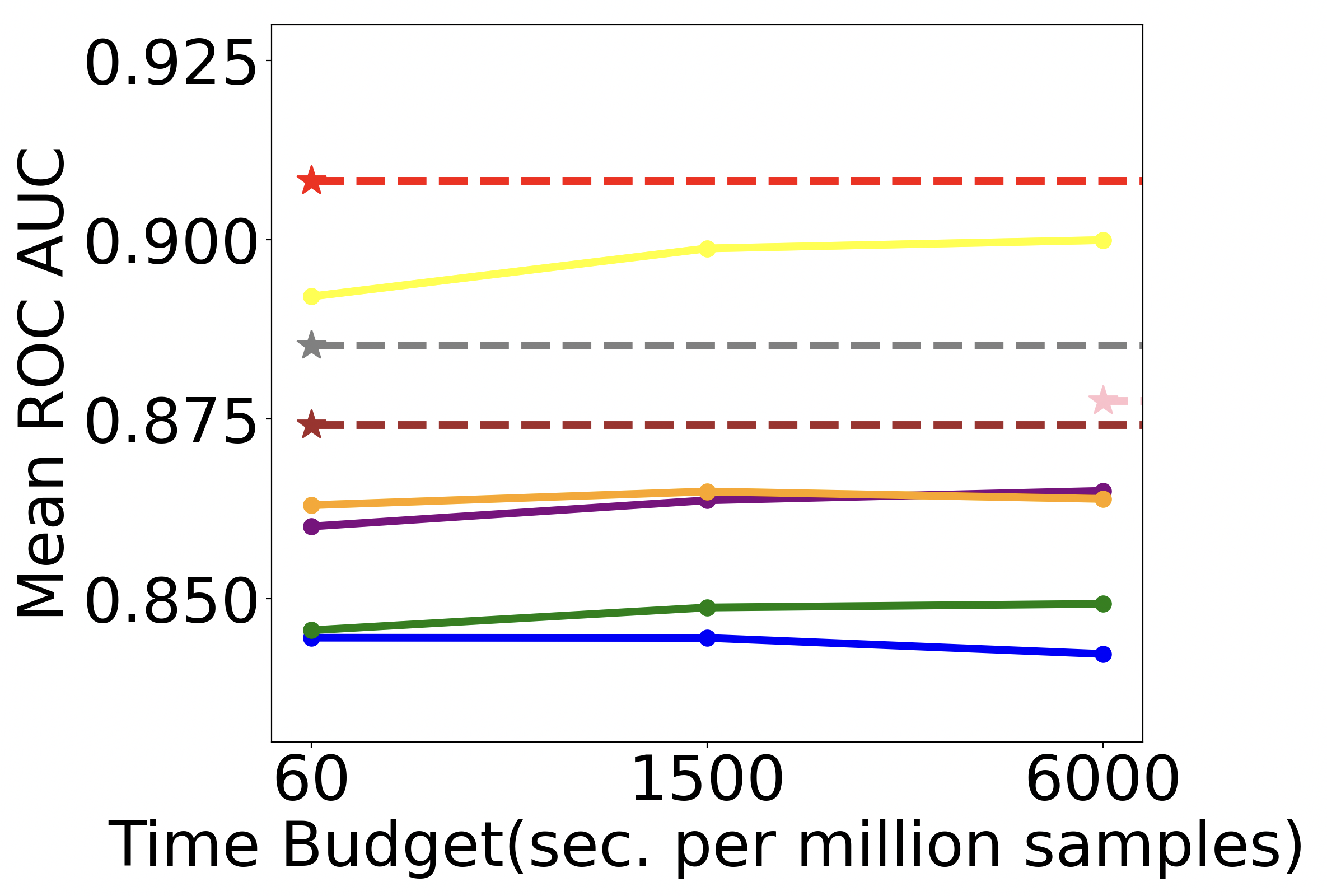}
\end{minipage}
\begin{minipage}[t]{0.32\textwidth}
\centering
\includegraphics[width=\textwidth]{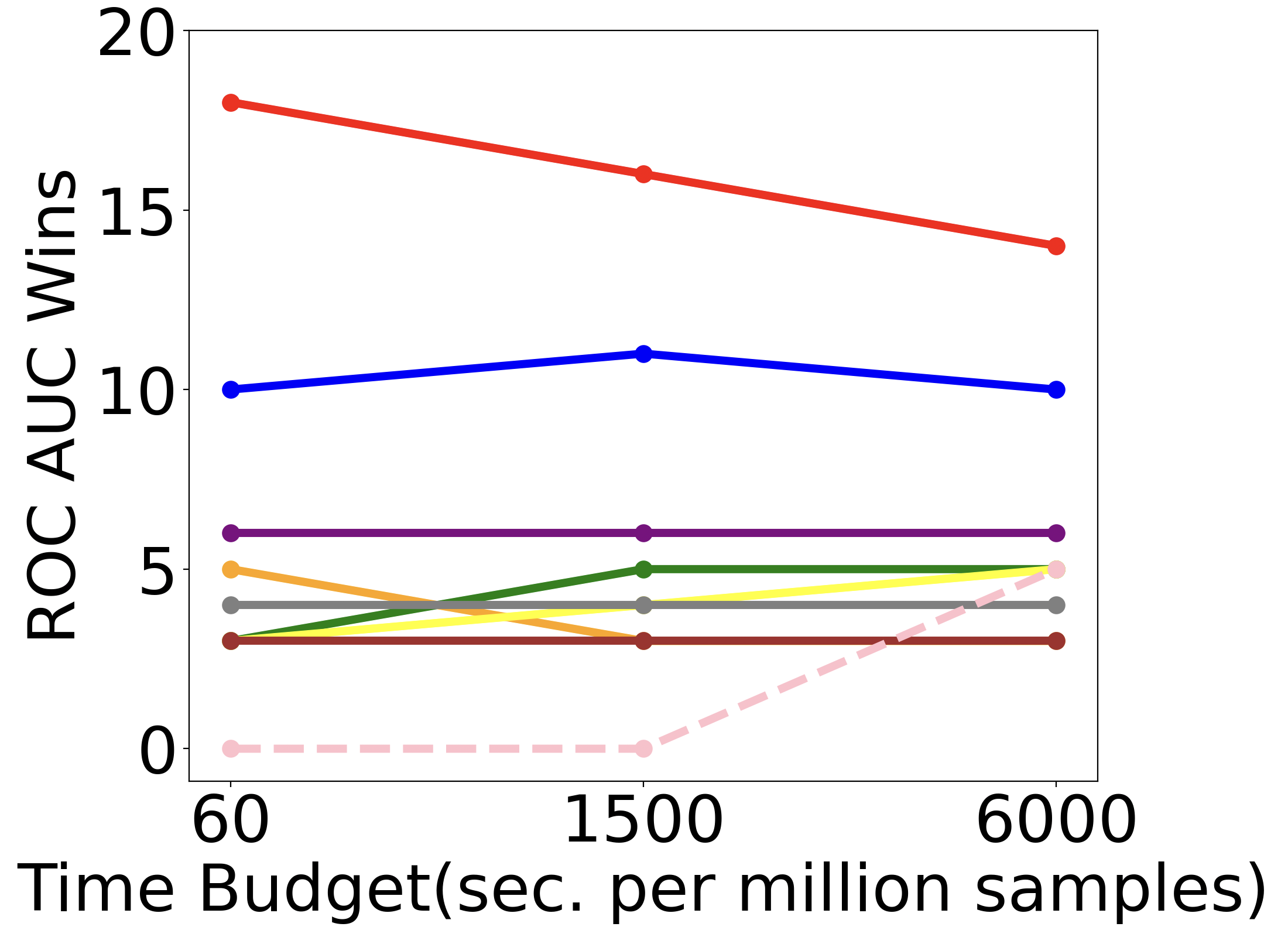}
\end{minipage}
\begin{minipage}[t]{0.32\textwidth}
\centering
\includegraphics[width=\textwidth]{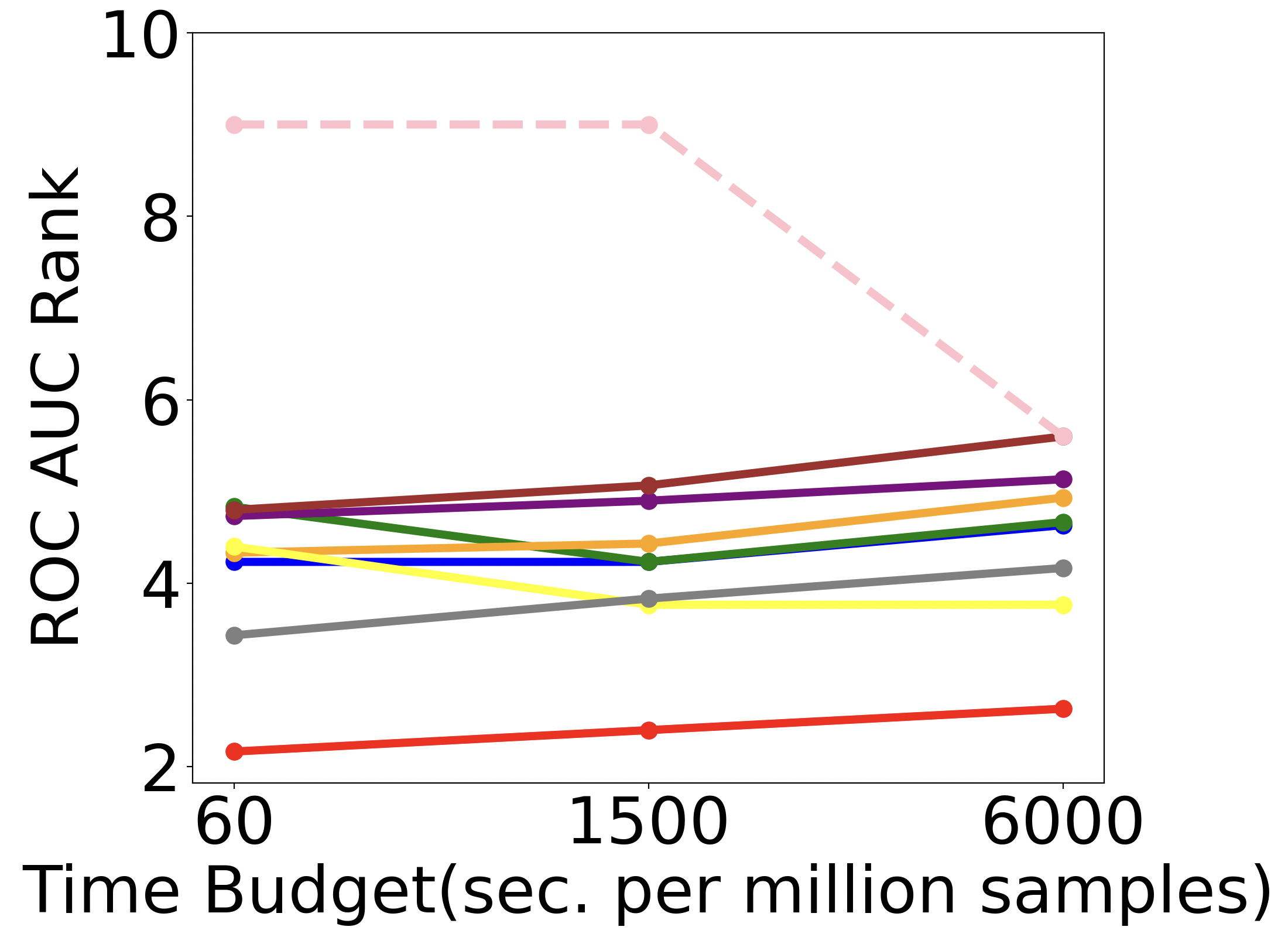}
\end{minipage}

\begin{minipage}[t]{0.6\textwidth}
\centering
\includegraphics[width=\textwidth]{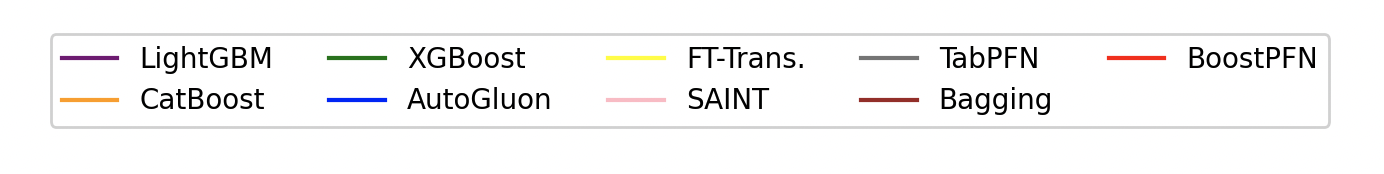}
\end{minipage}
    \label{fig:timelimitationfigures}
   
\end{figure*}

\textbf{Efficiency.} Next we show the performances of BoostPFN and baselines as the function of time limitation in Figure \ref{fig:timelimitationfigures}. All models use 5,000 training samples and details are shown in Supplemental Section \ref{appendix:timebudget}. The results show that BoostPFN can get best results when still maintaining efficiency, using much shorter time compared to GBDTs, deep learning methods and Autogluon, and a bit time longer than TabPFN and Bagging (While not shown in the figure, TabPFN costs about 12s and Bagging costs about 40s and BoostPFN costs about 60s per million samples). When more time is allowed, the results for baselines can be improved a bit.

\vspace*{-1mm}
\section{Ablations}
\vspace*{-1mm}
\subsection{Emperical Validation of Assumption 
\ref{prop:fittingassumption}}
\vspace*{-1mm}
\label{sec:addsampleablation}
We conduct the experiments based on TabPFN and on the dataset "analcatdata\_dmft" that it performs not well~\cite{hollmann2023tabpfn}. We examine the performances of TabPFN when modifying the input training set.

\textbf{Test Set.} We first aim to determine the upper bound of TabPFN on the test set. The results are shown in Table \ref{tab:fittingassumption}. $D_{train}+D_{train}$ refers to the concatenation of two identical training sets. $D_{train}+D_{test}$ refers to the concatenation of the training set and the test set. $D_{train}+2D_{test}$ refers to the concatenation of the training set and two identical test sets. The results show that adding the test set as part of the input training set for TabPFN significantly improves its performance, with the upper bound being the results of using $D_{test}$ as the sole input training set. Duplicating the test set does not help TabPFN perform better on the test set, indicating that this is indeed the upper bound for this test set. This upper bound suggests some interference between test samples that could decrease performance, supporting Assumption \ref{prop:fittingassumption}. We further validate the assumption through experiments on a single test sample.

\begin{table}[htbp]
    \centering
\vspace*{-1mm}
    \caption{Upper Bound of TabPFN on full test set. }
    
    \resizebox{0.5\textwidth}{!}{\begin{tabular}{c|ccc}
    \toprule
         Input Train set & $D_{train}+D_{train}$& $D_{train}+D_{test}$& $D_{train}+2D_{test}$\\
         \midrule
         AUC OVO & 0.5815 & 0.6812& 0.7135\\
         \midrule
         Input Train set & $D_{test}$&$D_{test}+D_{test}$\\
         \midrule
         AUC OVO & 0.7154 &0.7154 \\
    \bottomrule
    \end{tabular}}
    \label{tab:fittingassumption}
    \vspace*{-1mm}
\end{table}

\textbf{Single Test Sample.} Next, we demonstrate how the prediction for a single test sample depends on the input training samples. In Table \ref{tab:fittingononesample}, we show the probability of TabPFN predicting the correct label for just one test sample, $(x_{t1}, y_{t1})$, from the same dataset. 

First, we consider adding duplicates of the target sample to the training set. For this single sample, the input training set does not significantly aid the prediction, as the probability (0.192) is slightly higher than random guessing (0.167 for a 6-class classification task). Adding just one duplicate of this sample to the input training set improves the probability slightly. Adding two duplicates further increases the probability to 0.22. Subsequently, adding 10 duplicates increases the probability to 0.42, and adding 20 duplicates increases it to 0.75. Finally, adding 50 duplicates raises the probability to 0.94, and 100 duplicates to 0.98.

We then examine the results of replacing training samples with duplicates of the target sample. Similar to the trend observed when adding duplicates, replacing more training samples with the target sample also improves prediction accuracy (from 0.192 to 0.992 as the number of replacements increases from 0 to 100). Furthermore, with the same number of duplicates, replacing training samples performs better than simply adding them. This is reasonable because the training set with added duplicates contains more irrelevant samples than the training set where samples are replaced, further validating Assumption \ref{prop:fittingassumption}.

\begin{table}[htbp]
    \centering
\vspace*{-1mm}
    \caption{Probability of TabPFN predicting the right class  on single test sample. We show the number of adding or replacing with the probability. }
   \resizebox{0.5\textwidth}{!}{ \begin{tabular}{c|ccccccc}
    \toprule
        \# of Adding & 0 & 1 & 2&10 & 20 & 50& 100 \\
         \midrule
         Prob. & 0.192 & 0.206 & 0.220 & 0.417& 0.748&	0.945 &	0.981 \\
         \midrule
       \# of Replacing & 0 & 1 & 2&10 & 20 & 50& 100\\
         \midrule
         Prob. & 0.192 & 0.206 & 0.223 & 0.460& 0.787&	0.975 &	0.992   \\
    \bottomrule
    \end{tabular}}
\vspace*{-1mm}
    \label{tab:fittingononesample}
\end{table}

In a nutshell, TabPFN has performance upper bounds on the entire test set, even when duplicates of the test set are used as input. This may be because the prior is not well-suited for this dataset. However, for a specific sample, adding duplicates can increase the probability of predicting the correct class to as much as 98\%, validating that adding the target sample improves performance, as stated in Assumption \ref{prop:fittingassumption}. Additionally, replacing training samples with the target sample also improves the probability, further supporting the notion that replacing irrelevant samples with the target sample enhances performance, as proposed in Assumption \ref{prop:fittingassumption}.

\subsection{Sampling Weight Updates} 
\label{sec:ablation-updatingmethods}

In Section \ref{sec:updatingmethods} we introduce three updating methods in the BoostPFN. Here we compare the results of those methods in Table \ref{tab:updatingmethod}. Exp., Hada. and Ada. stands for ExpHadamard, Hadamard and Adaboost updating respectively. The OVO AUC results are very close across different updating methods, showing the robustness towards them. However, in practice we encourage choosing the best one to get best results.
\begin{table}[h]
    \centering
        \caption{Mean AUC OVO of three updating method in BoostPFN. }
\resizebox{0.3\textwidth}{!}{\begin{tabular}{c|ccc}

    \toprule
          \# of Training & Exp. & Hada.& Ada.\\
         <5000 & 0.893 & 0.890& 0.894 \\
         5000 & 0.900 & 0.898& 0.905 \\
         50000 & 0.905 & 0.907& 0.907 \\
         >50000 & 0.910 & 0.911& 0.907 \\
    \bottomrule
    \end{tabular}}\label{tab:updatingmethod}
\end{table}

\subsection{Emperical Convergence of BoostPFN}
\label{sec:convergenceemperical}
\begin{figure*}[htbp]
\centering
\caption{The first row shows boost loss on the training set of 3 datasets. The second row shows boost loss on the test set. The x-axis is the number of weak learners. From left to right, dataset order is BNG(primary-tumor), pokerhand, KDDCup99.}
\begin{minipage}[t]{0.3\textwidth}
\centering
\includegraphics[width=\textwidth]{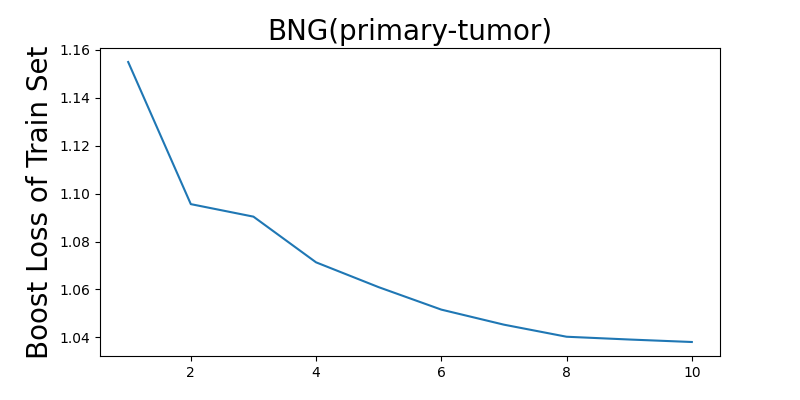}
\end{minipage}
\begin{minipage}[t]{0.3\textwidth}
\centering
\includegraphics[width=\textwidth]{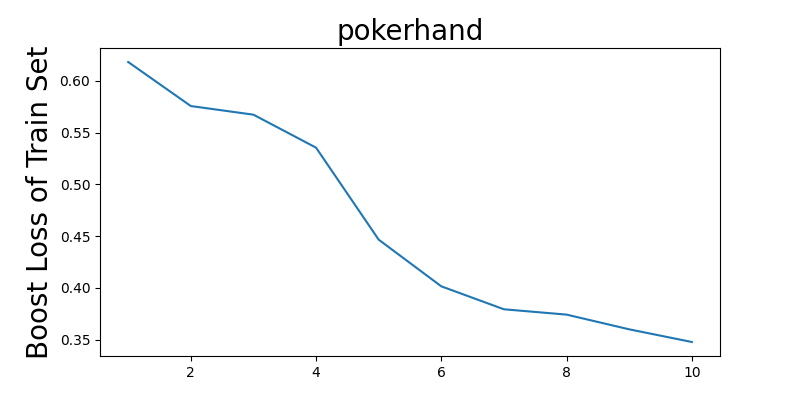}
\end{minipage}
\begin{minipage}[t]{0.3\textwidth}
\centering
\includegraphics[width=\textwidth]{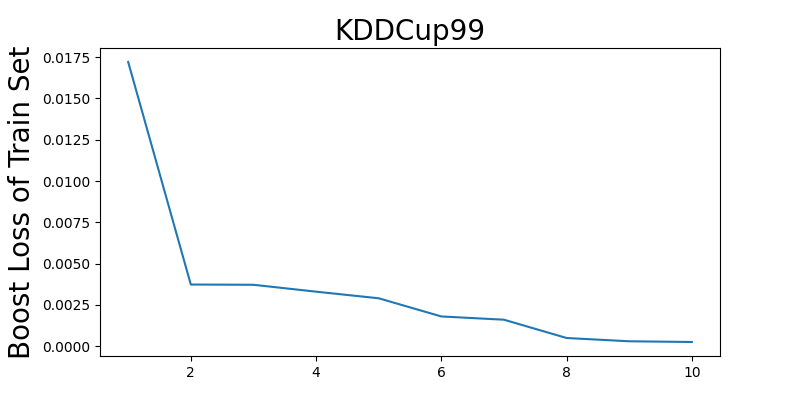}
\end{minipage}

\begin{minipage}[t]{0.3\textwidth}
\centering
\includegraphics[width=\textwidth]{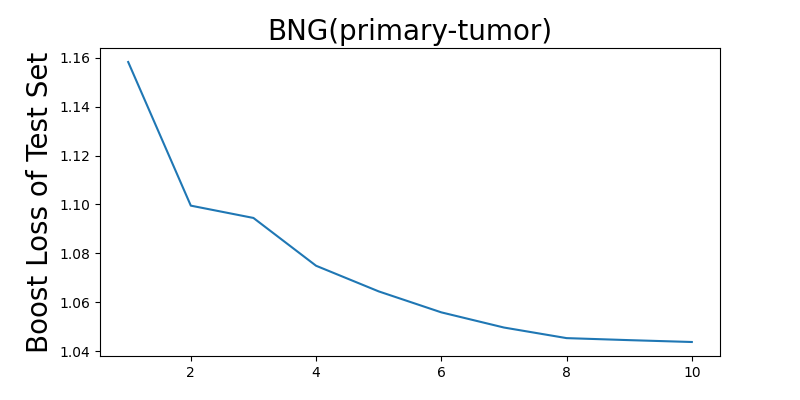}
\end{minipage}
\begin{minipage}[t]{0.3\textwidth}
\centering
\includegraphics[width=\textwidth]{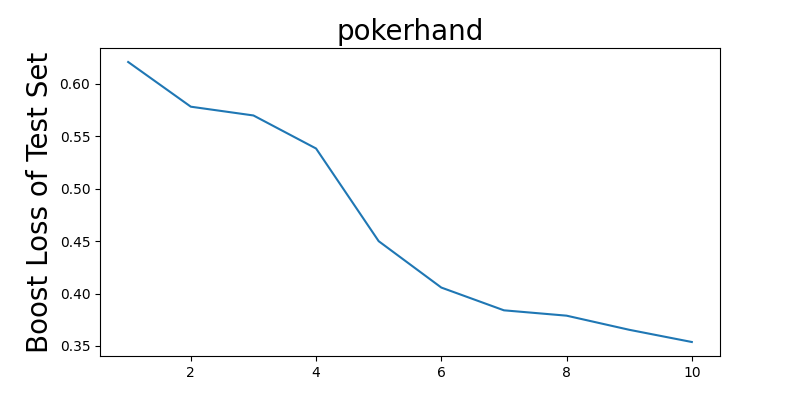}
\end{minipage}
\begin{minipage}[t]{0.3\textwidth}
\centering
\includegraphics[width=\textwidth]{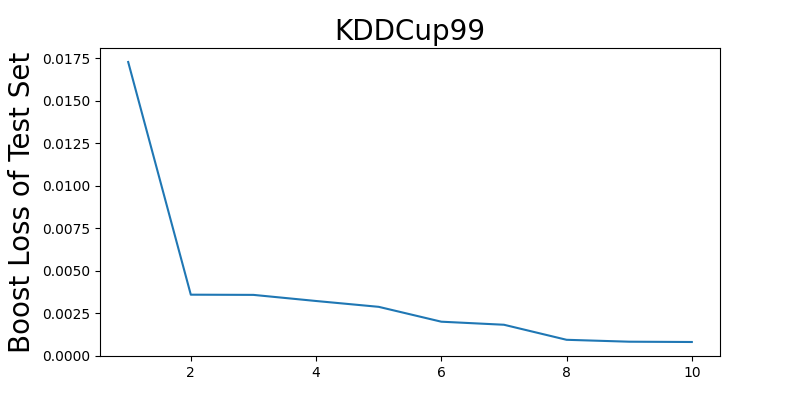}
\end{minipage}

\label{fig:boostprogressfor3data}

\end{figure*}

Although the fitting assumption of PFN is validated by experiments, we are curious about how the gradient boost actually works in large datasets. So we examine the boost loss on training set and test set and boost residual we used for gradient boost on training set. We use ExpHadamard Updating for the experiments. Full results of all datasets are showed in Appendix \ref{appendix:boostprogress}, while we show the results of three datasets (BNG(primary-tumor), pokerhand, KDDCup99) in Figure \ref{fig:boostprogressfor3data}, on which BoostPFN improves the best compared to TabPFN. We can find that when the number of weak learners increase, the boost loss on training set and test set decrease at the same time, showing the effectiveness of our method.

\vspace{-2mm}
\section{Related Works}
\vspace{-2mm}
\textbf{Tabular Data Classification.}
Machine Learning methods and Deep Learning methods are two main types of method for tabular data classification. For Machine learning methods, variants of GBDTs~\cite{friedman2001greedy} dominates this field mainly because of their fast training time and robustness~\cite{shwartzziv2021tabular}. LightGBM~\cite{ke2017lightgbm}, CatBoost~\cite{prokhorenkova2018catboost} and XGBoost~\cite{chen2016xgboost} are three popular GBDTs applied in tabular data classification. Previous works have found that deep learning models are not better than the performance of GBDT or AutoML methods for small to medium-sized tabular data (< 10,000 samples; while matching GBDT performance on larger datasets)~\cite{borisov2022deep,grinsztajn2022tree,shwartz2022tabular}. Furthermore, deep learning methods often use custom parameter tuning and preprocessing that makes the training on large datasets much more time-consuming, including FT-Transformer, SAINT, etc.~\cite{gorishniy2021revisiting,somepalli2021saint,arik2021tabnet,kadra2021well,kossen2021self}. There are also AutoML systems like AutoSklearn2~\cite{feurer2022auto} and Autogluon~\cite{erickson2020autogluon}. AutoSklearn2 uses Bayesian Optimization and combines the evaluated models into a weighted ensemble and Autogluon combines a broad kind of models including neural networks and tree-based models into a stacked ensemble.

\textbf{PFNs.} PFNs are proposed by \cite{muller2021transformers} on binary classification and small tabular datas of 100 samples. \cite{hollmann2023tabpfn} further improve it to 10-class classification and medium-size (thousands of samples) tabular datas. ~\cite{pmlr-v202-nagler23a} analyze the statistical foundation of a single PFN and not an ensemble of PFNs. Larger datasets remain un-discovered in this field and  our work extends the field of PFNs to larger datasets.

\textbf{Ensemble for Transformers.}
Ensemble methods like AdaBoost~\cite{hastie2009multi} or Gradient Boosting~\cite{friedman2001greedy,friedman2002stochastic} are commonly applied on decision trees for better performances. Recently, Transformers are combined with AdaBoost in many fields. \cite{hou2023promptboosting} integrate LLM into AdaBoost for natural language inference (NLI) tasks, by training an MLP projection of the final hidden state of a special token. \cite{manikandan2023language} concentrates on condensing knowledge
into an intermediary representation referred to as "summary." and applies AdaBoost for In-Context Learning for the prediction. However, their combination of summary sampling and AdaBoost are not theoretically validated. In this paper we propose a more generalized Gradient boosting method that is well validated theoretically and performs better in experiments.

\vspace{-2mm}
\section{Conclusion}
\vspace{-2mm}
In this paper, we have introduced BoostPFN, an innovative approach to scaling Prior-Fitted Networks (PFNs) for larger datasets, addressing the scalability challenges encountered by PFNs when applied to extensive data sets. Our method, through a meticulous investigation into the fitting assumptions of PFNs and the selection of input samples, presents a gradient boosting framework that significantly enhances the performance of PFNs, particularly on a large-scale. BoostPFN significantly extends the data size that PFNs can effectively process—up to 50 times the pre-training dataset size—while maintaining high performance compared to GBDTs, deep learning methods, and AutoML systems. We also provide a theoretical analysis that substantiates its convergence. BoostPFN offers a new perspective on handling large-scale tabular data classification tasks efficiently. In the future when PFNs are trained on larger datasets and perform better (for example, \cite{hollmann2025accurate}, which is published while our paper is under review), BoostPFN can still boost the predictions on \textit{much} larger datasets with the new, better PFNs.

\section{Acknowledgement}

No acknowledgement to report at this time..


\bibliography{tabular,transfomer}
\bibliographystyle{plain}
\section*{Checklist}

 \begin{enumerate}

 \item For all models and algorithms presented, check if you include:
 \begin{enumerate}
   \item A clear description of the mathematical setting, assumptions, algorithm, and/or model. [Yes]
   \item An analysis of the properties and complexity (time, space, sample size) of any algorithm. [Yes]
   \item (Optional) Anonymized source code, with specification of all dependencies, including external libraries. [Not Applicable]
 \end{enumerate}

 \item For any theoretical claim, check if you include:
 \begin{enumerate}
   \item Statements of the full set of assumptions of all theoretical results. [Yes]
   \item Complete proofs of all theoretical results. [Yes]
   \item Clear explanations of any assumptions. [Yes]     
 \end{enumerate}

 \item For all figures and tables that present empirical results, check if you include:
 \begin{enumerate}
   \item The code, data, and instructions needed to reproduce the main experimental results (either in the supplemental material or as a URL). [Yes]
   \item All the training details (e.g., data splits, hyperparameters, how they were chosen). [Yes]
         \item A clear definition of the specific measure or statistics and error bars (e.g., with respect to the random seed after running experiments multiple times). [Yes]
         \item A description of the computing infrastructure used. (e.g., type of GPUs, internal cluster, or cloud provider). [Yes]
 \end{enumerate}

 \item If you are using existing assets (e.g., code, data, models) or curating/releasing new assets, check if you include:
 \begin{enumerate}
   \item Citations of the creator If your work uses existing assets. [Yes]
   \item The license information of the assets, if applicable. [Yes]
   \item New assets either in the supplemental material or as a URL, if applicable. [Yes]
   \item Information about consent from data providers/curators. [Yes]
   \item Discussion of sensible content if applicable, e.g., personally identifiable information or offensive content. [Not Applicable]
 \end{enumerate}

 \item If you used crowdsourcing or conducted research with human subjects, check if you include:
 \begin{enumerate}
   \item The full text of instructions given to participants and screenshots. [Not Applicable]
   \item Descriptions of potential participant risks, with links to Institutional Review Board (IRB) approvals if applicable. [Not Applicable]
   \item The estimated hourly wage paid to participants and the total amount spent on participant compensation. [Not Applicable]
 \end{enumerate}

 \end{enumerate}

\input{appendix}

\end{document}

%% file: appendix.tex
\newpage
\appendix
\onecolumn
\section{Additional Theoretical Analysis}
\label{appendix:proof}
\subsection{A simplified Scenerio for Dataset Optimization}
\label{appendix:proofforlemma}
We consider a simplified scenario in which we need to update the sampling weights from a uniform distribution.
\begin{lemma}
\label{lemma:improveprob} Given a training set $D$, $N=|D|$ is the number of training samples. We do the non-replacement sampling in $D$ with the sampling size $z$ and uniform sampling weights $w_1=(p_1^{w_1},p_2^{w_1},...p_N^{w_1})$, where each $p_a^{w_1}=p_0=\frac{1}{N}$ for $a \in (1,2,...,N)$. Then with a new set of sampling weights $w_2$ where $p_i^{w_2}=kp_i^{w_1}$ and $p_j^{w_2}=\lambda p_j^{w_1}$ for all $j \in (1,2,...,N) $ and $ j\neq i$, where $k<N$ and $\lambda=\frac{1-kp_0}{1-p_0}$ to maintain normalization, given Assumption \ref{prop:fittingassumption} and for $k>(N-z)(1-\lambda_B)+1$, the expectation of predicting the right label on target $i$ is larger than with $w_1$ : 
\begin{align}
\label{eq:lemma1}
    \mathbb{E}[q_{(\theta,w_2)}(y_i|x_i,D)]> \mathbb{E}[q_{(\theta,w_1)}(y_i|x_i,D)],
\end{align}
where $0<\lambda_B<1$ and is defined in Eq. \ref{eq:definitionlambdaB}.
\end{lemma}
We put the proof below.
\begin{proof}
First we write the expectation in 
\begin{align}
    \label{eq:expectationoflemma}
    \mathbb{E}[q_{(\theta,w)}(y_i|x_i,D)]=\sum_{D_w^z \in \mathcal{D}^z} p (D_w^z)q_{\theta}(y_i|x_i,D_w^z),
\end{align}
where $\mathcal{D}^z$ is the space of all possibilities of non-replacement sampling with size $z$.

Now we consider the situation that $D^z=D_{0} \cup (x_i,y_i)$, then there are $(N-z)$ situations that $\bar{D}^z=D_{0} \cup (x_j,y_j)$ where $(x_j,y_j) \notin D^z$. So the expectation can be written as 
\begin{align}
    \label{eq:expectationout}
    \mathbb{E}[q_{(\theta,w)}(y_i|x_i,D)]=\sum_{D^z}\Big[ p (D^z;w)q_{\theta}(y_i|x_i,D^z) + \sum_{\bar{D}^z} p (\bar{D}^z;w)q_{\theta}(y_i|x_i,\bar{D}^z)\Big].
\end{align}
Let $A(w)=p (D^z;w)q_{\theta}(y_i|x_i,D^z)$ and $B(w)=\sum_{\bar{D}^z} p (\bar{D}_w^z;w)q_{\theta}(y_i|x_i,\bar{D}^z)$. To prove Eq. \ref{eq:lemma1}, we need to show that $A(w_2)+B(w_2)>A(w_1)+B(w_1)$.

In the non-replacement sampling, if we use $p_a$ to denote the $a$th sample's weight, $A(w_1)$ can be written as 
\begin{align}
    A(w_1)&=\frac{z! \cdot \prod_{a=1}^z p_a}{\prod_{a=1}^z (1 - \sum_{b=1}^{a-1} p_b)} q_{\theta}(y_i|x_i,D^z),\\\nonumber
    &=\frac{z!}{\binom{N}{z}} q_{\theta}(y_i|x_i,D^z) ,\text{  remembering each weight is $p_0=\frac{1}{N}$ for $w_1$}.
\end{align}
Then we compute $A(w_2)$, remembering that in $w_2$, $p_i^{w_2}=kp_0$ and for $a\neq i$, $p_a^{w_2}=\lambda p_0 =\frac{1-kp_0}{1-p_0} p_0$,
\begin{align}
     A(w_2)&=\frac{z! \cdot p_i \cdot \prod_{a=1}^{z-1} p_a^{w_2}}{\prod_{a=1}^z (1 - \sum_{b=1}^{a-1} p_b^{w_2})} q_{\theta}(y_i|x_i,D^z),\\\nonumber
     &=\frac{z! \cdot \left(k \cdot \frac{1}{N}\right) \cdot \left(\lambda \cdot \frac{1}{N}\right)^{z-1}}{\frac{N \cdot (N-k) \cdot (N-k-\lambda) \cdots (N - k - (z-2)\lambda)}{N^z}}q_{\theta}(y_i|x_i,D^z)\\\nonumber
     &=k \cdot \lambda^{z-1} \cdot \frac{z! \cdot N^z}{N \cdot (N-k) \cdot (N-k-\lambda) \cdots (N - k - (z-2)\lambda)}q_{\theta}(y_i|x_i,D^z)
\end{align}
So 
\begin{align}
    \frac{A(w_2)}{A(w_1)}&=\frac{k \cdot \lambda^{z-1} \cdot \frac{z! \cdot N^z}{N \cdot (N-k) \cdot (N-k-\lambda) \cdots (N - k - (z-2)\lambda)}}{\frac{z!}{\binom{N}{z}}} \\\nonumber
   & = k \cdot \lambda^{z-1} \cdot N^z \cdot \frac{\binom{N}{z}}{N \cdot (N-k) \cdot (N-k-\lambda) \cdots (N - k - (z-2)\lambda)} \\\nonumber
   &  = k \cdot \lambda^{z-1} \cdot \frac{N \cdot (N-1) \cdot (N-2) \cdots (N-(z-1))}{N \cdot (N-k) \cdot (N-k-\lambda) \cdots (N - k - (z-2)\lambda)}
\end{align}
Note that 
\begin{align}
    \lambda&=\frac{1-kp_0}{1-p_0}=\frac{1-k\frac{1}{N}}{1-\frac{1}{N}}=\frac{N-k}{N-1},
\end{align}
so 
\begin{align}
    \frac{N-1}{N-k}=\frac{N-2}{N-k-\lambda}=\cdots=\frac{N-(z-1)}{N-k-(z-2)\lambda}=\frac{1}{\lambda}.
\end{align}
Then 
\begin{align}
    \frac{A(w_2)}{A(w_1)}=k \cdot \lambda^{z-1} \cdot \lambda^{1-z}=k.
\end{align}
Using the same process we can compute that 
\begin{align}
    \frac{B(w_2)}{B(w_1)}&=\lambda^z\frac{\prod_{a=1}^z (1 - \sum_{b=1}^{a-1} p_0)}{\prod_{a=1}^z (1 - \sum_{b=1}^{a-1} \lambda p_0)} .
\end{align}
We then define 
\begin{align}
    \label{eq:definitionlambdaB}
    \lambda_B \triangleq \lambda^z\frac{\prod_{a=1}^z (1 - \sum_{b=1}^{a-1} p_0)}{\prod_{a=1}^z (1 - \sum_{b=1}^{a-1} \lambda p_0)}.
\end{align}
With $0<\lambda<1$ we know $0<\lambda_B<1$. So 
\begin{align}
\label{eq:A+B>A+B}
    A(w_2)+B(w_2)>A(w_1)+B(w_1) &\Longleftrightarrow  kA(w_1)+\lambda_BB(w_1)>A(w_1)+B(w_1)  \\\nonumber
    &\Longleftrightarrow  (k-1)A(w_1)>(1-\lambda_B)B(w_1) \\\nonumber
    &\Longleftrightarrow \frac{A(w_1)}{B(w_1)} > \frac{1-\lambda_B}{k-1}.
\end{align}
Remembering the definition of $A(w_1)$ and $B(w_1)$, we know that 
\begin{align}
    \frac{A(w_1)}{B(w_1)} = \frac{ q_{\theta}(y_i|x_i,D^z)}{(N-z) q_{\theta}(y_i|x_i,\bar{D}^z)}> \frac{1}{(N-z)}, \text{~~~ Given Assumption \ref{prop:fittingassumption}.}
\end{align}
So Eq. \ref{eq:A+B>A+B} holds if 
\begin{align}
     \frac{1}{(N-z)}>\frac{1-\lambda_B}{k-1}.
\end{align}
This is true because $k>(N-z)(1-\lambda_B)+1$ is one of our condition. Then we complete the proof of Lemma \ref{lemma:improveprob}.
\end{proof}
\subsection{Proof for Theorem \ref{theorem:RGBM}}
\label{appendix:prooffortheorem}
We begin the proof by the first proving that BoostPFN is a kind of Randomized Gradient Boosting Machine.
\begin{remark}
    BoostPFN showed in Algorithm \ref{alg:BoostPFN} is a kind of Randomized Gradient Boosting Machine.
\end{remark}
\begin{proof}
Randomized Gradient Boosting Machine can be written as 

\begin{algorithm}[h]
\caption{Randomized Gradient Boosting Machine (RGBM)}\label{al:rgbm}

\begin{algorithmic}
\STATE {\bf Initialization.}  Initialize with $f^{0}(x)=0$.\\

For $m=0,\ldots,M-1$ do:\\


\ \ \ (1) Compute pseudo-residual $r^m=-\left[\frac{\partial \ell(y_{i},f^{m}(x_{i}))}{\partial f^m(x_{i})}\right]_{i=1,\ldots,n}.$

\ \ \ (2) Pick a random subset $J$ of weak-learners by \emph{some rule} (i.e., one of Type 0 - Type 3)

\ \ \ (3) Find the best weak-learner in $J$: $j_{m} =\text{argmin}_{j\in J}\min_{\sigma} \sum_{i=1}^n (r^m_i-\sigma b(x_i;\tau_j))^{2}$.

\ \ \ (4) Choose the step-size $\rho_m$ by one of the following rules:

\ \ \ \ \ \ $ \bullet $ line-search: $\rho_{m}=\text{argmin}_{\rho}\sum_{i=1}^{n}\ell(y_{i},f^{m}(x_{i})+\rho b(x_i;\tau_{j_m}))$;

\ \ \ \ \ \ $ \bullet $ constant step-size: $\rho_m=\rho \left(\sum_{i=1}^n r_i^m b(x_i;\tau_{j_m})\right)$, where $\rho$ is a constant specified a priori.

\ \ \ (5) Update the model $f^{m+1}(x)=f^{m}(x)+\rho_{m}b(x;\tau_{j_{m}})$.\\

\smallskip

\STATE  {\bf Output.}  $f^{M}(x)$.

\end{algorithmic}
\end{algorithm}
from \cite{lu2020randomized}. $b(x_i;\tau_{j_m})$ is the weak learner in round $m$. Type 0 to Type 3 are as follows :
\begin{itemize}
    \item[] {\bf [Type 0]:} {\it(Full Deterministic Selection)} We choose $J$ as the whole set of weak-learners. This is a deterministic selection rule.
    
    \item[] {\bf [Type 1]:} {\it(Random Selection)} We choose 
    uniformly at random $t$ weak-learners from all possible weak-learners without replacement---the collection is denoted by $J$.
    
    \item[] {\bf [Type 2]:} {\it(Random Single Group Selection)} Given a non-overlapping partition of the weak-learners, we pick one group uniformly at random and denote the collection of weak-learners in that group by $J$. 
    
    \item[] {\bf [Type 3]:} {\it(Random Multiple Group Selection)} Given a non-overlapping partition of the weak-learners, we pick $t$ groups uniformly at random and let the collection of weak-learners across these groups be $J$.
\end{itemize}
When we compare RGBM with BoostPFN, it's clear that if we choose line-search in step (4) of RGBM, the difference is in step (2) and (3). Here we show that the weights updating is actually picking random subset of weak learner. 

We write the subset of TabPFN as $J=\{q_{\theta}(\cdot|x,D_w^z)|D_w \text{ generated by } w\}$. In BoostPFN algorithm, we sample a weak learner from the subset instead of choosing the best one. So we need to modify the algorithm a bit here. We introduce the union of subset with different sampling weights $J_U=\bigcup_{i=0,1,2} J_i=\{q_{\theta}(\cdot|x,D_{w_i}^z)|D_{w_i}^z \text{ generated by } w_i \}$ where $w_0$ is uniform sampling weights, $w_1$ is the sampling weights for this round of boost, $w_2$ is the sampling weights for last round. And we choose the best weak learner in this subset. The subset is picked via Type 3. Thus BoostPFN with this modification is the same as RGBM\footnote{This modification does not improve the empirical results too much but costs about 3x times, so considering the time cost we don't use this modification in our experiments}.
\end{proof}
Then the Theorem 4.2 in \cite{lu2020randomized} that the RGBM in Algorithm 2 if $\ell$ is a $\sigma$-smooth function and
has a bounded level set will converge with the rate of $O(\frac{\sigma}{M})$ will lead straightforwardly to the proof of Eq. \ref{eq:theorem}.

\section{Experimental Implementation and Hyperparameters}
\label{appendix:experimentalimplementation}
We do all experiments on a platform with 48 CPU cores and RTX 3090. For ensemble models on small datasets, we use sampling size 0.5 of the training set if training set contains less than 1000 samples. The hyperparameter spaces for LightGBM, CatBoost and XGBoost follow \cite{hollmann2023tabpfn} and are shown in Table \ref{app:tab:spaces}.

\begin{table}[htbp]
\caption{Hyperparameter spaces for baselines.}
\label{app:tab:spaces}
\centering
\begin{tabular}{lllll}
\toprule
baseline &  name & type & log & range \\
\midrule
\multirow{10}{*}{LightGBM} & num\_leaves & int & [5, 50] & yes \\
    & max\_depth & int & [3, 20] & yes \\
    & learning\_rate & float & [$e^{-3}$, 1] & - \\
    & n\_estimators & int & 50, 2000 & - \\
    & min\_child\_weight & float & [$e^{-5}$, $e^4$] & yes \\
    & reg\_alpha & float & [0, 1e-1, 1, 2, 5, 7, 10, 50, 100] & yes \\
    & reg\_lambda & float & [0, 1e-1, 1, 5, 10, 20, 50, 100] & yes \\
    & subsample & float & [0.2, 0.8] & - \\
    \midrule
\multirow{6}{*}{CatBoost} & learning\_rate & float & [$e^{-5}$, 1] & yes \\
    & random\_strength & int & [1, 20] & - \\
    & l2\_leaf\_reg & float & [1, 10] & yes \\
    & bagging\_temperature & float & [0, 1.0] & yes \\
    & leaf\_estimation\_iterations & int & [1, 20] & - \\
    & iterations & int & [100, 4000] & - \\
\midrule
\multirow{10}{*}{XGBoost} & learning\_rate & float & [$e^{-7}$, 1] & yes \\
    & max\_depth & int & [1, 10] & - \\
    & subsample & float & [0.2, 1] & - \\
    & colsample\_bytree & float & [0.2, 1] & - \\
    & colsample\_bylevel & float & [0.2, 1] & - \\
    & min\_child\_weight & float & [$e^{-16}$, $e^5$] & yes \\
    & alpha & float & [$e^{-16}$, $e^2$] & yes \\
    & lambda & float & [$e^{-16}$, $e^2$] & yes \\
    & gamma & float & [$e^{-16}$, $e^2$] & yes \\
    & n\_estimators & int & [100, 4000] & - \\

\bottomrule
\end{tabular}
\end{table}
For AutoGluon we use "best quality". For FT-Transformer we tuning learning rate in [5e-5,1e-4,5e-4,1e-3], batch size in [128, 256, 512, 1024]. For SAINT we use default setting because it costs too much time to tuning.

For tuning methods, time limitation is 6000 seconds per million samples.

\section{Datasets Statistics}
\label{appendix:datasets-statistics}
We show the dataset statistics for small datasets in Table \ref{tab:smalltest_datasets_table}, cited from \cite{hollmann2023tabpfn}. Large datasets statistics are shown in Table \ref{tab:largetest_datasets_table}.
\begin{table}[htbp]
    \caption{Small datasets used for the evaluation. All 30 datasets are at most $2\,000$ samples, $100$ features and $10$ classes.}
    \label{tab:smalltest_datasets_table}
    \centering
    \tiny
    \begin{tabular}{@{\hskip 0mm}l@{\hskip 0mm}rrrrrrr}
\toprule
 Name & \#Feat. &  \#Cat. &  \#Inst. &  \#Class. &  \#NaNs &  Minor. Class Size & OpenML ID \\
\midrule
 balance-scale & 5 & 1 & 625 & 3 &   0 & 49 &    11 \\
 mfeat-fourier & 77 & 1 & 2000 & 10 &   0 & 200 &    14 \\
 breast-w & 10 & 1 & 699 & 2 &  16 & 241 &    15 \\
 mfeat-karhunen & 65 & 1 & 2000 & 10 &   0 & 200 &    16 \\
 mfeat-morphological & 7 & 1 & 2000 & 10 &   0 & 200 &    18 \\
 mfeat-zernike & 48 & 1 & 2000 & 10 &   0 & 200 &    22 \\
 cmc & 10 & 8 & 1473 & 3 &   0 & 333 &    23 \\
 credit-approval & 16 &    10 & 690 & 2 &  67 & 307 &    29 \\
 credit-g & 21 &    14 & 1000 & 2 &   0 & 300 &    31 \\
 diabetes & 9 & 1 & 768 & 2 &   0 & 268 &    37 \\
 tic-tac-toe & 10 &    10 & 958 & 2 &   0 & 332 &    50 \\
 vehicle & 19 & 1 & 846 & 4 &   0 & 199 &    54 \\
 eucalyptus & 20 & 6 & 736 & 5 & 448 & 105 &   188 \\
 analcatdata\_auth... & 71 & 1 & 841 & 4 &   0 & 55 &   458 \\
 analcatdata\_dmft & 5 & 5 & 797 & 6 &   0 & 123 &   469 \\
 pc4 & 38 & 1 & 1458 & 2 &   0 & 178 &  1049 \\
 pc3 & 38 & 1 & 1563 & 2 &   0 & 160 &  1050 \\
 kc2 & 22 & 1 & 522 & 2 &   0 & 107 &  1063 \\
 pc1 & 22 & 1 & 1109 & 2 &   0 & 77 &  1068 \\
 banknote-authenti... & 5 & 1 & 1372 & 2 &   0 & 610 &  1462 \\
blood-transfusion-... & 5 & 1 & 748 & 2 &   0 & 178 &  1464 \\
ilpd & 11 & 2 & 583 & 2 &   0 & 167 &  1480 \\
qsar-biodeg & 42 & 1 & 1055 & 2 &   0 & 356 &  1494 \\
wdbc & 31 & 1 & 569 & 2 &   0 & 212 &  1510 \\
cylinder-bands & 40 &    22 & 540 & 2 & 999 & 228 &  6332 \\
dresses-sales & 13 &    12 & 500 & 2 & 835 & 210 & 23381 \\
MiceProtein & 82 & 5 & 1080 & 8 &    1396 & 105 & 40966 \\
car & 7 & 7 & 1728 & 4 &   0 & 65 & 40975 \\
steel-plates-fault & 28 & 1 & 1941 & 7 &   0 & 55 & 40982 \\
climate-model-simu... & 21 & 1 & 540 & 2 &   0 & 46 & 40994 \\
\bottomrule
\end{tabular}

\end{table}

\begin{table}[htbp]
    \caption{Large datasets used for the evaluation. All 30 datasets are at most 10 million samples, 100 features and 144 classes.}
    \label{tab:largetest_datasets_table}
    \centering
    \tiny
\begin{tabular}{@{\hskip 0mm}l@{\hskip 0mm}rrrrrrr}
\toprule
 Name & \#Feat. &  \#Cat. &  \#Inst. &  \#Class. &  \#NaNs &  Minor. Class Size & OpenML ID \\
\midrule
BNG(page-blocks,nominal,295245) & 11 & 11 & 295245 & 5 & 0 & 1558 & 125 \\
BNG(glass,nominal,137781) & 10 & 10 & 137781 & 7 & 0 & 307 & 133 \\
BNG(heart-c,nominal,1000000) & 14 & 14 & 1000000 & 5 & 0 & 1618 & 136 \\
BNG(heart-h,nominal,1000000) & 14 & 14 & 1000000 & 5 & 0 & 1659 & 138 \\
BNG(waveform-5000,nominal,1000000) & 41 & 41 & 1000000 & 3 & 0 & 330548 & 147 \\
pokerhand & 11 & 6 & 829201 & 10 & 0 & 2 & 155 \\
RandomRBF\_0\_0 & 11 & 1 & 1000000 & 5 & 0 & 92713 & 156 \\
RandomRBF\_10\_1E-3 & 11 & 1 & 1000000 & 5 & 0 & 92713 & 157 \\
RandomRBF\_10\_1E-4 & 11 & 1 & 1000000 & 5 & 0 & 92713 & 158 \\
SEA(50) & 4 & 1 & 1000000 & 2 & 0 & 385658 & 161 \\
SEA(50000) & 4 & 1 & 1000000 & 2 & 0 & 385668 & 162 \\
BNG(heart-c) & 14 & 8 & 1000000 & 5 & 0 & 1609 & 266 \\
BNG(primary-tumor) & 18 & 18 & 1000000 & 22 & 0 & 1417 & 1177 \\
BNG(solar-flare) & 13 & 13 & 1000000 & 3 & 0 & 1393 & 1179 \\
Stagger1 & 4 & 4 & 1000000 & 2 & 0 & 111609 & 1236 \\
Stagger2 & 4 & 4 & 1000000 & 2 & 0 & 444057 & 1237 \\
Stagger3 & 4 & 4 & 1000000 & 2 & 0 & 333571 & 1238 \\
AirlinesCodrnaAdult & 30 & 17 & 1076790 & 2 & 7275 & 473652 & 1240 \\
skin-segmentation & 4 & 1 & 245057 & 2 & 0 & 50859 & 1502 \\
creditcard & 31 & 1 & 284807 & 2 & 0 & 492 & 1597 \\
BNG(spambase) & 58 & 58 & 1000000 & 2 & 0 & 394052 & 40515 \\
BNG(anneal) & 39 & 33 & 1000000 & 6 & 0 & 555 & 40520 \\
fars & 30 & 16 & 100968 & 8 & 0 & 9 & 40672 \\
seattlecrime6 & 8 & 6 & 523590 & 144 & 6916 & 1 & 41960 \\
porto-seguro & 38 & 26 & 595212 & 2 & 846458 & 21694 & 42206 \\
CreditCardFraudDetection & 31 & 1 & 284807 & 2 & 0 & 492 & 42397 \\
KDDCup99 & 42 & 10 & 4898431 & 23 & 0 & 2 & 42746 \\
bates\_classif\_20 & 21 & 1 & 5100000 & 2 & 0 & 2549577 & 45654 \\
colon & 63 & 1 & 5100000 & 2 & 0 & 2549437 & 45665 \\
breast & 78 & 1 & 5100000 & 2 & 0 & 2549502 & 45669 \\

 \bottomrule
\end{tabular}

\end{table}
\section{Full Experiment Results}
\label{appendix:fullexperiments}
We show the per dataset results on small datasets with a 1 hour time limit in Table \ref{tab:fullresultssmall}. The large datasets results with different number of training samples are showed in Table \ref{tab:fullresultslarge-5000},\ref{tab:fullresultslarge-50000} and \ref{tab:fullresultslarge-full} .
\begin{table}[htbp]
    \centering
    \caption{Per dataset results on small datasets lower than 5000 training samples.}
    \resizebox{\textwidth}{!}{\begin{tabular}{@{\hskip 0mm}
l@{\hskip 1mm}
l@{\hskip 1mm}
l@{\hskip 1mm}
l@{\hskip 1mm}
l@{\hskip 1mm}
l@{\hskip 1mm}
l@{\hskip 1mm}
l@{\hskip 1mm}|
l@{\hskip 1mm}
l@{\hskip 0mm}}
\toprule
{} 	&	        LightGBM 	&	        CatBoost 	&	          XGBoost 	&	               AutoGluon 	&	 FT-Trans.	&	 SAINT 	&	  TabPFN 	&	           Bagging 	&	           BoostPFN  	\\
\midrule
balance-scale	&	0.9938	&	0.9245	&	0.9939	&	0.9919	&	0.9935	&	0.86366	&	0.9973	&	0.9985	&	0.9996	\\
mfeat-fourier	&	0.9786	&	0.9816	&	0.9803	&	0.9843	&	0.9782	&	0.978938	&	0.9811	&	0.9761	&	0.9769	\\
breast-w	&	0.991	&	0.9931	&	0.9896	&	0.9933	&	0.9846	&	0.987477	&	0.9934	&	0.9922	&	0.9921	\\
mfeat-karhunen	&	0.9979	&	0.9986	&	0.9983	&	0.9987	&	0.9961	&	0.998036	&	0.9978	&	0.9981	&	0.999	\\
mfeat-morphologica..	&	0.9601	&	0.9629	&	0.9612	&	0.9698	&	0.9665	&	0.95501	&	0.9669	&	0.9669	&	0.9664	\\
mfeat-zernike	&	0.9716	&	0.9759	&	0.9735	&	0.9908	&	0.9808	&	0.973376	&	0.9823	&	0.9834	&	0.9833	\\
cmc	&	0.7288	&	0.7256	&	0.7299	&	0.7331	&	0.7073	&	0.699643	&	0.7276	&	0.7173	&	0.7205	\\
credit-approval	&	0.9415	&	0.9389	&	0.9422	&	0.9415	&	0.9175	&	0.933137	&	0.9322	&	0.9424	&	0.9427	\\
credit-g	&	0.7684	&	0.7852	&	0.7853	&	0.7941	&	0.7644	&	0.602838	&	0.7894	&	0.7916	&	0.7918	\\
diabetes	&	0.8247	&	0.8383	&	0.8378	&	0.8391	&	0.8475	&	0.650567	&	0.841	&	0.8251	&	0.8239	\\
tic-tac-toe	&	0.9988	&	0.9992	&	1	&	1	&	0.9935	&	0.562474	&	0.9759	&	0.9437	&	0.9767	\\
vehicle	&	0.9232	&	0.9302	&	0.9282	&	0.9416	&	0.9357	&	0.923496	&	0.9589	&	0.9563	&	0.9597	\\
eucalyptus	&	0.8931	&	0.8979	&	0.9004	&	0.9204	&	0.8961	&	0.851701	&	0.9245	&	0.9196	&	0.9227	\\
analcatdata\_author..	&	0.9999	&	0.9999	&	0.9997	&	0.9993	&	0.9972	&	0.999019	&	1	&	1	&	1	\\
analcatdata\_dmft	&	0.5461	&	0.5589	&	0.5743	&	0.5657	&	0.5489	&	0.551028	&	0.579	&	0.5811	&	0.5806	\\
pc4	&	0.9301	&	0.9413	&	0.9291	&	0.9428	&	0.9254	&	0.914273	&	0.9383	&	0.9207	&	0.9247	\\
pc3	&	0.8178	&	0.8247	&	0.8288	&	0.8282	&	0.7911	&	0.808226	&	0.8373	&	0.8513	&	0.8518	\\
kc2	&	0.8141	&	0.8323	&	0.8227	&	0.8242	&	0.8059	&	0.846287	&	0.8346	&	0.8705	&	0.8698	\\
pc1	&	0.8321	&	0.86	&	0.8489	&	0.8578	&	0.7207	&	0.805705	&	0.8761	&	0.8936	&	0.8876	\\
banknote-authentic..	&	1	&	1	&	1	&	1	&	0.9927	&	0.992147	&	1	&	1	&	1	\\
blood-transfusion-..	&	0.7144	&	0.7403	&	0.7312	&	0.7364	&	0.7803	&	0.928475	&	0.7549	&	0.7747	&	0.7687	\\
ilpd	&	0.6917	&	0.7279	&	0.7171	&	0.723	&	0.7044	&	0.619734	&	0.7379	&	0.7523	&	0.7527	\\
qsar-biodeg	&	0.9126	&	0.9217	&	0.9191	&	0.9276	&	0.9202	&	0.999062	&	0.9336	&	0.9282	&	0.9288	\\
wdbc	&	0.9904	&	0.9931	&	0.9904	&	0.9956	&	0.986	&	0.709422	&	0.9964	&	0.9985	&	0.9988	\\
cylinder-bands	&	0.8556	&	0.8757	&	0.8782	&	0.8878	&	0.8038	&	0.790092	&	0.8336	&	0.7802	&	0.7969	\\
dresses-sales	&	0.5593	&	0.5696	&	0.5823	&	0.5507	&	0.5056	&	0.578095	&	0.5376	&	0.5559	&	0.5532	\\
MiceProtein	&	0.9997	&	0.9999	&	0.9998	&	1	&	0.9992	&	0.999702	&	0.9999	&	0.9998	&	1	\\
car	&	0.9925	&	0.9955	&	0.9948	&	0.997	&	0.9849	&	0.953275	&	0.995	&	0.9902	&	0.9965	\\
steel-plates-fault..	&	0.9626	&	0.9655	&	0.9656	&	0.9666	&	0.9532	&	0.59391	&	0.9655	&	0.9595	&	0.9587	\\
climate-model-simu..	&	0.9286	&	0.9344	&	0.9255	&	0.9391	&	0.8719	&	0.666173	&	0.9415	&	0.926	&	0.9298	\\

\midrule

Mean AUC OVO
                                         &  0.884$\pm$.012 &   0.89$\pm$.011 &   0.891$\pm$.011 &  \textbf{0.895}$\pm$.011    &  0.875$\pm$.010       &  0.824$\pm$.011     & 0.894$\pm$.010 &  0.893$\pm$.010&\textbf{0.895}$\pm$.009 \\
                                         
\bottomrule
\end{tabular}}
    
    \label{tab:fullresultssmall}
\end{table}

\begin{table}[htbp]
    \centering
        \caption{Per dataset results on large datasets with 5,000 training samples.}
    \label{tab:fullresultslarge-5000}
    \resizebox{\textwidth}{!}{\begin{tabular}{l|lllllll|ll}
\toprule
{} 	&	        LightGBM 	&	        CatBoost 	&	          XGBoost 	&	               AutoGluon 	&	 FT-Trans.	&	 SAINT 	&	  TabPFN 	&	           Bagging 	&	           BoostPFN  	\\
\midrule
BNG(page-blocks,nominal,295245)	&	0.8401	&	0.8385	&	0.8296	&	0.7812	&	0.8478	&	0.7531	&	0.81	&	0.7851	&	0.8473	\\
BNG(glass,nominal,137781)	&	0.9011	&	0.8979	&	0.8994	&	0.8322	&	0.9013	&	0.9007	&	0.8907	&	0.8395	&	0.8824	\\
BNG(heart-c,nominal,1000000)	&	0.7664	&	0.7779	&	0.7825	&	0.6283	&	0.8019	&	0.7876	&	0.7856	&	0.7754	&	0.8072	\\
BNG(heart-h,nominal,1000000)	&	0.829	&	0.7803	&	0.793	&	0.6898	&	0.8156	&	0.7867	&	0.8001	&	0.786	&	0.8188	\\
BNG(waveform-5000,nominal,1000000)	&	0.9569	&	0.9567	&	0.9572	&	0.9579	&	0.9541	&	0.9568	&	0.9536	&	0.9535	&	0.9536	\\
pokerhand	&	0.636	&	0.6512	&	0.6205	&	0.6134	&	0.9395	&	0.9011	&	0.6894	&	0.7725	&	0.9632	\\
RandomRBF\_0\_0	&	0.9878	&	0.9884	&	0.9886	&	0.9927	&	0.9872	&	0.9898	&	0.9915	&	0.9902	&	0.9918	\\
RandomRBF\_10\_1E-3	&	0.9554	&	0.9591	&	0.9592	&	0.9671	&	0.9554	&	0.9598	&	0.9623	&	0.9607	&	0.9629	\\
RandomRBF\_10\_1E-4	&	0.9635	&	0.9667	&	0.9659	&	0.9758	&	0.9627	&	0.9654	&	0.9727	&	0.9675	&	0.9701	\\
SEA(50)	&	0.8728	&	0.8768	&	0.8761	&	0.8781	&	0.8765	&	0.8688	&	0.8776	&	0.8777	&	0.8773	\\
SEA(50000)	&	0.8736	&	0.8754	&	0.8755	&	0.8781	&	0.8771	&	0.8803	&	0.8769	&	0.8774	&	0.8765	\\
BNG(heart-c)	&	0.7768	&	0.7712	&	0.7676	&	0.727	&	0.7564	&	0.7594	&	0.7618	&	0.7514	&	0.7598	\\
BNG(primary-tumor)	&	0.9009	&	0.8915	&	0.8977	&	0.8558	&	0.891	&	0.8473	&	0.8347	&	0.8443	&	0.8908	\\
BNG(solar-flare)	&	0.8783	&	0.8673	&	0.8698	&	0.8527	&	0.8628	&	0.8303	&	0.8448	&	0.8323	&	0.8867	\\
Stagger1	&	1	&	1	&	1	&	1	&	1	&	1	&	1	&	1	&	1	\\
Stagger2	&	1	&	1	&	1	&	1	&	1	&	1	&	1	&	1	&	1	\\
Stagger3	&	1	&	1	&	1	&	1	&	1	&	1	&	1	&	1	&	1	\\
AirlinesCodrnaAdult	&	0.8738	&	0.8761	&	0.8756	&	0.8822	&	0.8753	&	0.8525	&	0.8702	&	0.8679	&	0.8708	\\
skin-segmentation	&	0.9997	&	0.9998	&	0.9998	&	1	&	0.9999	&	0.9985	&	0.9999	&	0.9995	&	1	\\
creditcard	&	0.947	&	0.9621	&	0.9616	&	0.961	&	0.9624	&	0.9435	&	0.9763	&	0.9601	&	0.9844	\\
BNG(spambase)	&	0.6452	&	0.6615	&	0.6659	&	0.6632	&	0.6618	&	0.3719	&	0.6596	&	0.6532	&	0.6588	\\
BNG(anneal)	&	0.9618	&	0.9622	&	0.9627	&	0.8851	&	0.9643	&	0.9491	&	0.952	&	0.9351	&	0.9624	\\
fars	&	0.5021	&	0.4896	&	0.4972	&	0.4969	&	0.8509	&	0.8552	&	0.8751	&	0.8059	&	0.8774	\\
seattlecrime6	&	0.988	&	0.9882	&	0.5022	&	0.8889	&	0.9902	&	0.9452	&	0.9879	&	0.9	&	0.9926	\\
porto-seguro	&	0.5742	&	0.5993	&	0.5983	&	0.5955	&	0.5941	&	0.5632	&	0.5919	&	0.5921	&	0.6086	\\
CreditCardFraudDetection	&	0.9732	&	0.9611	&	0.9556	&	0.9466	&	0.9706	&	0.9355	&	0.9699	&	0.9698	&	0.9828	\\
KDDCup99	&	0.5196	&	0.5248	&	0.5181	&	0.5248	&	0.835	&	0.8876	&	0.7609	&	0.674	&	0.9535	\\
bates\_classif\_20	&	0.8701	&	0.8689	&	0.8727	&	0.8737	&	0.8754	&	0.8684	&	0.8774	&	0.8771	&	0.8761	\\
colon	&	0.9939	&	0.9936	&	0.9949	&	0.9967	&	0.9969	&	0.9976	&	0.996	&	0.9916	&	0.9972	\\
breast	&	0.9626	&	0.9613	&	0.975	&	0.9903	&	0.9925	&	0.9715	&	0.9895	&	0.9862	&	0.9934	\\

\midrule
Mean AUC OVO              &0.865$\pm$.001  &        0.865$\pm$.001 &          0.849$\pm$.001 &                        0.844$\pm$.001 &  0.900$\pm$.002 &           0.878$\pm$.002 &           0.885$\pm$.001 &0.874$\pm$.001 &\textbf{0.908}$\pm$.001\\
\bottomrule
\end{tabular}}

\end{table}

\newpage

\begin{table}[htbp]
    \centering
        \caption{Per dataset results on large datasets with 50,000 training samples.}
    \label{tab:fullresultslarge-50000}
  \resizebox{\textwidth}{!}{  \begin{tabular}{l|lllllll|ll}
\toprule
{} 	&	        LightGBM 	&	        CatBoost 	&	          XGBoost 	&	               AutoGluon 	&	 FT-Trans.	&	 SAINT 	&	  TabPFN 	&	           Bagging 	&	           BoostPFN  	\\
\midrule
BNG(page-blocks,nominal,295245)	&	0.854	&	0.8575	&	0.8518	&	0.8415	&	0.8575	&	0.8445	&	OOM	&	0.8209	&	0.8427	\\
BNG(glass,nominal,137781)	&	0.9151	&	0.9091	&	0.9155	&	0.9129	&	0.916	&	0.9126	&	OOM	&	0.8936	&	0.8941	\\
BNG(heart-c,nominal,1000000)	&	0.8125	&	0.7999	&	0.8066	&	0.8102	&	0.811	&	0.808	&	OOM	&	0.7767	&	0.802	\\
BNG(heart-h,nominal,1000000)	&	0.8198	&	0.8042	&	0.813	&	0.8137	&	0.8095	&	0.8161	&	OOM	&	0.7872	&	0.8193	\\
BNG(waveform-5000,nominal,1000000)	&	0.9631	&	0.9619	&	0.9636	&	0.9583	&	0.9616	&	0.9626	&	OOM	&	0.9548	&	0.9536	\\
pokerhand	&	0.4945	&	0.5275	&	0.4923	&	0.4981	&	0.9625	&	0.8915	&	OOM	&	0.7813	&	0.9752	\\
RandomRBF\_0\_0	&	0.9929	&	0.9902	&	0.9916	&	0.993	&	0.9935	&	0.994	&	OOM	&	0.9909	&	0.9924	\\
RandomRBF\_10\_1E-3	&	0.9649	&	0.9627	&	0.9673	&	0.9685	&	0.9701	&	0.9678	&	OOM	&	0.9634	&	0.964	\\
RandomRBF\_10\_1E-4	&	0.9767	&	0.9718	&	0.9774	&	0.9826	&	0.9827	&	0.9806	&	OOM	&	0.9697	&	0.9745	\\
SEA(50)	&	0.8817	&	0.8811	&	0.8808	&	0.89	&	0.8775	&	0.8731	&	OOM	&	0.8783	&	0.8767	\\
SEA(50000)	&	0.8806	&	0.8805	&	0.8805	&	0.8892	&	0.877	&	0.8729	&	OOM	&	0.8779	&	0.8767	\\
BNG(heart-c)	&	0.7977	&	0.77	&	0.7828	&	0.7911	&	0.7924	&	0.7641	&	OOM	&	0.7633	&	0.7684	\\
BNG(primary-tumor)	&	0.9104	&	0.9163	&	0.9134	&	0.8695	&	0.9138	&	0.9072	&	OOM	&	0.8548	&	0.8901	\\
BNG(solar-flare)	&	0.9121	&	0.9001	&	0.9061	&	0.8808	&	0.9018	&	0.8796	&	OOM	&	0.8416	&	0.8772	\\
Stagger1	&	1	&	1	&	1	&	1	&	1	&	1	&	OOM	&	1	&	1	\\
Stagger2	&	1	&	1	&	1	&	1	&	1	&	1	&	OOM	&	1	&	1	\\
Stagger3	&	1	&	1	&	1	&	1	&	1	&	1	&	OOM	&	1	&	1	\\
AirlinesCodrnaAdult	&	0.8921	&	0.8932	&	0.5006	&	0.8967	&	0.8915	&	0.8937	&	OOM	&	0.8709	&	0.8751	\\
skin-segmentation	&	0.9999	&	0.9999	&	1	&	1	&	0.9999	&	0.9998	&	OOM	&	0.9999	&	1	\\
creditcard	&	0.9756	&	0.974	&	0.9721	&	0.9781	&	0.9719	&	0.9492	&	OOM	&	0.9683	&	0.9833	\\
BNG(spambase)	&	0.6688	&	0.6695	&	0.669	&	0.6693	&	0.6689	&	0.661	&	OOM	&	0.6614	&	0.6577	\\
BNG(anneal)	&	0.98	&	0.9818	&	0.9778	&	0.9875	&	0.9884	&	0.9856	&	OOM	&	0.9372	&	0.9712	\\
fars	&	0.8866	&	0.8867	&	0.9086	&	0.8566	&	0.8937	&	0.9003	&	OOM	&	0.859	&	0.8768	\\
seattlecrime6	&	0.9898	&	0.5105	&	0.5042	&	0.5041	&	0.9891	&	0.9637	&	OOM	&	0.9906	&	0.9913	\\
porto-seguro	&	0.6084	&	0.5675	&	0.6034	&	0.6257	&	0.6192	&	0.6153	&	OOM	&	0.6141	&	0.6136	\\
CreditCardFraudDetection	&	0.9766	&	0.9757	&	0.9742	&	0.9779	&	0.9639	&	0.9656	&	OOM	&	0.9703	&	0.9824	\\
KDDCup99	&	0.4863	&	0.4869	&	0.5352	&	0.4974	&	0.8298	&	0.9276	&	OOM	&	0.7649	&	0.9606	\\
bates\_classif\_20	&	0.8746	&	0.8744	&	0.874	&	0.8782	&	0.8779	&	0.8905	&	OOM	&	0.8784	&	0.8775	\\
colon	&	0.9965	&	0.9965	&	0.9962	&	0.9973	&	0.9976	&	1	&	OOM	&	0.9915	&	0.9974	\\
breast	&	0.9826	&	0.9749	&	0.9844	&	0.9776	&	0.9942	&	0.9904	&	OOM	&	0.9898	&	0.9942	\\

\midrule
Mean AUC OVO &0.883$\pm$.001 &        0.864$\pm$.001 &         0.855$\pm$.001 &    0.865$\pm$.001&\textbf{0.910}$\pm$.001& 0.907$\pm$.001        &  OOM &           0.8888$\pm$.001 &           \textbf{0.910}$\pm$.001 \\
\bottomrule
\end{tabular}}

\end{table}
\newpage

\begin{table}[htbp]
    \centering
        \caption{Per dataset results on large datasets with full training samples.}
    \label{tab:fullresultslarge-full}
    \resizebox{\textwidth}{!}{\begin{tabular}{l|lllllll|ll}
\toprule
{} 	&	        LightGBM 	&	        CatBoost 	&	          XGBoost 	&	               AutoGluon 	&	 FT-Trans.	&	 SAINT 	&	  TabPFN 	&	           Bagging 	&	           BoostPFN  	\\
\midrule
BNG(page-blocks,nominal,295245)	&	0.8587	&	0.8602	&	0.8598	&	0.8456	&	0.8619	&	0.8472	&	OOM	&	0.8149	&	0.8471	\\
BNG(glass,nominal,137781)	&	0.9154	&	0.9148	&	0.9092	&	0.9129	&	0.918	&	0.9152	&	OOM	&	0.8936	&	0.9	\\
BNG(heart-c,nominal,1000000)	&	0.8137	&	0.8148	&	0.8105	&	0.8104	&	0.8196	&	0.817	&	OOM	&	0.777	&	0.8052	\\
BNG(heart-h,nominal,1000000)	&	0.8205	&	0.8221	&	0.8216	&	0.822	&	0.8246	&	0.8194	&	OOM	&	0.7877	&	0.8138	\\
BNG(waveform-5000,nominal,1000000)	&	0.9663	&	0.9667	&	0.9659	&	0.9649	&	0.9654	&	0.9643	&	OOM	&	0.9551	&	0.9561	\\
pokerhand	&	0.8174	&	0.8751	&	0.8782	&	0.8968	&	0.9721	&	0.9555	&	OOM	&	0.8096	&	0.968	\\
RandomRBF\_0\_0	&	0.9948	&	0.9934	&	0.9925	&	0.9951	&	0.9959	&	0.9959	&	OOM	&	0.9909	&	0.9929	\\
RandomRBF\_10\_1E-3	&	0.9761	&	0.9706	&	0.9678	&	0.981	&	0.9814	&	0.9811	&	OOM	&	0.9636	&	0.9658	\\
RandomRBF\_10\_1E-4	&	0.9866	&	0.9804	&	0.9771	&	0.9884	&	0.99	&	0.9904	&	OOM	&	0.97	&	0.975	\\
SEA(50)	&	0.9083	&	0.901	&	0.9162	&	0.9824	&	0.8767	&	0.8779	&	OOM	&	0.8782	&	0.8784	\\
SEA(50000)	&	0.9047	&	0.899	&	0.892	&	0.9824	&	0.8782	&	0.8777	&	OOM	&	0.878	&	0.878	\\
BNG(heart-c)	&	0.8013	&	0.7987	&	0.7935	&	0.7955	&	0.7998	&	0.7995	&	OOM	&	0.7649	&	0.779	\\
BNG(primary-tumor)	&	0.9188	&	0.9203	&	0.9181	&	0.9174	&	0.918	&	0.9139	&	OOM	&	0.8557	&	0.902	\\
BNG(solar-flare)	&	0.9262	&	0.9235	&	0.9207	&	0.9046	&	0.9346	&	0.928	&	OOM	&	0.8454	&	0.8946	\\
Stagger1	&	1	&	1	&	1	&	1	&	1	&	1	&	OOM	&	1	&	1	\\
Stagger2	&	1	&	1	&	1	&	1	&	1	&	1	&	OOM	&	1	&	1	\\
Stagger3	&	1	&	1	&	1	&	1	&	1	&	1	&	OOM	&	1	&	1	\\
AirlinesCodrnaAdult	&	0.9102	&	0.9035	&	0.9024	&	0.9134	&	0.9038	&	0.9029	&	OOM	&	0.8713	&	0.8822	\\
skin-segmentation	&	1	&	1	&	1	&	1	&	1	&	0.9997	&	OOM	&	0.9999	&	1	\\
creditcard	&	0.9799	&	0.9802	&	0.9802	&	0.9834	&	0.9778	&	0.9777	&	OOM	&	0.9712	&	0.982	\\
BNG(spambase)	&	0.6723	&	0.6722	&	0.672	&	0.6717	&	0.6714	&	0.6701	&	OOM	&	0.6609	&	0.6653	\\
BNG(anneal)	&	0.9949	&	0.9946	&	0.9939	&	0.995	&	0.9961	&	0.9953	&	OOM	&	0.9361	&	0.982	\\
fars	&	0.8769	&	0.9177	&	0.917	&	0.8428	&	0.8809	&	0.9128	&	OOM	&	0.874	&	0.8894	\\
seattlecrime6	&	0.99	&	0.9906	&	0.9912	&	0.9925	&	0.9915	&	0.9668	&	OOM	&	0.991	&	0.9911	\\
porto-seguro	&	0.6362	&	0.6362	&	0.6278	&	0.64	&	0.6311	&	0.6333	&	OOM	&	0.6163	&	0.6218	\\
CreditCardFraudDetection	&	0.9775	&	0.9832	&	0.9802	&	0.9839	&	0.9449	&	0.9779	&	OOM	&	0.9772	&	0.9821	\\
KDDCup99	&	0.926	&	0.7513	&	0.9509	&	0.7862	&	0.9434	&	0.9543	&	OOM	&	0.9035	&	0.9654	\\
bates\_classif\_20	&	0.8785	&	0.8785	&	0.8766	&	0.8789	&	0.8791	&	0.8787	&	OOM	&	0.8785	&	0.8778	\\
colon	&	0.9976	&	0.9976	&	0.9968	&	0.9973	&	0.9978	&	0.9977	&	OOM	&	0.9915	&	0.9975	\\
breast	&	0.9925	&	0.9936	&	0.9756	&	0.9745	&	0.9948	&	0.9946	&	OOM	&	0.9901	&	0.9944	\\

\midrule
Mean AUC OVO & 0.915$\pm$.001 &        0.911$\pm$.001 &          0.916$\pm$.001 &             0.915$\pm$.001 & \textbf{0.918}$\pm$.001 & \textbf{0.918}$\pm$.001&  OOM &           0.895$\pm$.001 &              0.913$\pm$.001 \\
\bottomrule
\end{tabular}}

\end{table}

\section{Additional Information for Time Budget}
\label{appendix:timebudget}

We follow the time budget comparison method from prior work (reference \cite{hollmann2023tabpfn}) and we can include further details in a revision. As for CPUs and GPUs, we show our hardware in the supplemental Section \ref{appendix:experimentalimplementation}; for reference here we use an Intel(R) Xeon(R) Server CPU with 48 cores with RTX 3090 GPU.

The overall training/inference time includes training across potentially multiple hyperparameter trials and inference, and represents the average cost of each model on all datasets. For each dataset, we compute overall time via $num_{trials}$ $\times$ $cost_\text{per trial}$, where $num_{trials}$ is chosen as the largest value such that ($num_{trials}$ - 1) $\times$ $cost_\text{per trial}$ < time budget, so the actual time cost  can be different across different models. We show the overall time for different models when granted different time budgets in the new table below (the chosen time budgets are drawn from Figure 2 of our submission). Note that SAINT only has results for budget 6000s/million samples because only one trial will cost more than the smaller time budgets listed in the table. We remark that when the budget is 60s/million samples, $num_{trials}$ = 1 for all non-TabPFN-based models.  Hence in such cases, the listed time cost reduces to the cost of a single training and inference run.

\begin{table}[htbp]
    \centering
        \caption{Per dataset results on large datasets with full training samples.}
    \label{tab:timebudget}
    \resizebox{\textwidth}{!}{\begin{tabular}{l|lllllll|ll}
\toprule
Time Budget per million samples (s)  	&	        LightGBM 	&	        CatBoost 	&	          XGBoost 	&	               AutoGluon 	&	 FT-Trans.	&	 SAINT 	&	  TabPFN 	&	           Bagging 	&	           BoostPFN  	\\
\midrule
60 &  128.7 & 755.2 &159.5 &613.7 &283.5&-& 11.1& 36.9 &49.7 \\
 1500 &  1192.1 & 2406.5 &1823.1 &2780.6 &1134.0 &-& - & - &-\\
 6000 &  3130.7 & 7439.3 &6349.2 &7229.8 &4536.0 &1545.7 & - & - &- \\
\bottomrule
\end{tabular}}

\end{table}

For the larger datasets, we use BoostPFN with different weak learners. The number of weak learners in the ensemble model is 10 for 5,000 training samples, 100 for 50,000 and 1,000 for full training set. For other models like XGBoost or LightGBM, the time limitation is still 6000 seconds per million samples.

\section{Boosting Process for Large Datasets}
\label{appendix:boostprogress}
\begin{figure}
    \centering
    \includegraphics[width=0.8\textwidth]{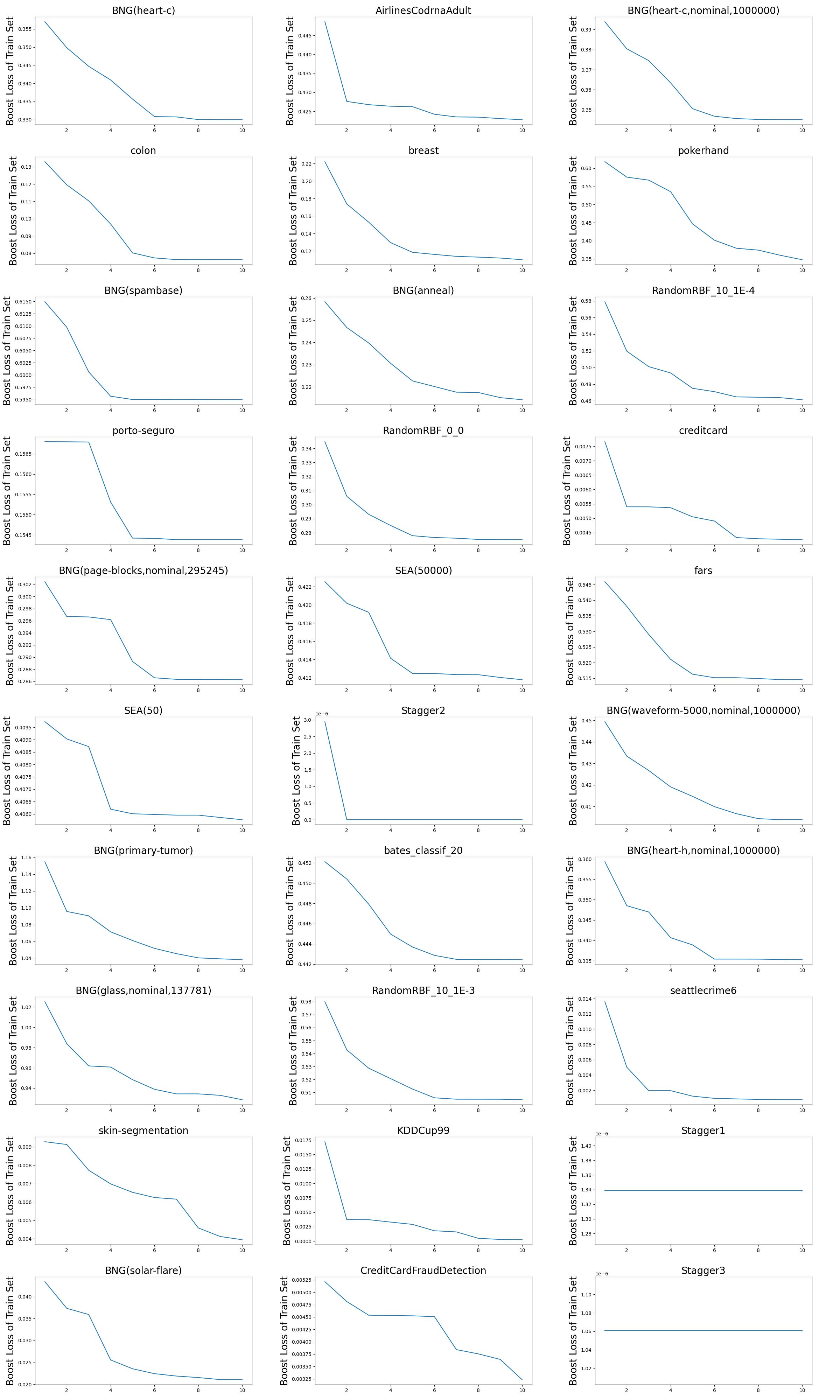}
    \caption{Boost loss on training set for all large datasets with 10 weak learners.}
    \label{fig:boost-lossfull}
\end{figure}

\begin{figure}
    \centering
    \includegraphics[width=0.8\textwidth]{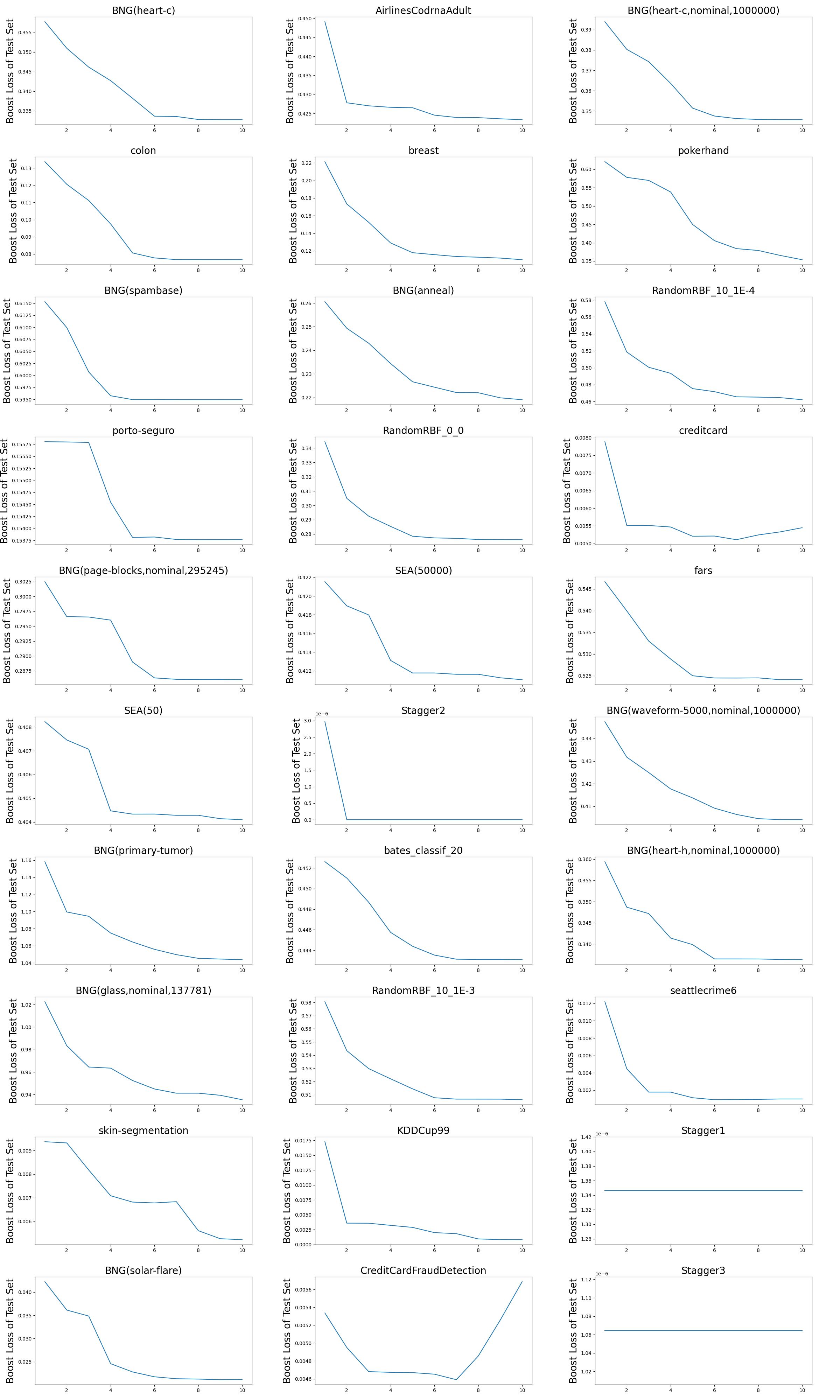}
    \caption{Boost loss on test set for all large datasets with 10 weak learners.}
    \label{fig:boost-loss-testfull}
\end{figure}

We show here the boosting loss on training set in Figure \ref{fig:boost-lossfull}, boosting loss on test set in Figure \ref{fig:boost-loss-testfull}. It's noted that in one of the datasets the test boosting loss goes up when the number of weak learners increase, while the training loss goes down, which clearly shows over-fitting.

